%% file: main.tex
\documentclass[10pt,journal,compsoc]{IEEEtran}

\usepackage{
amsfonts,
cite,
pifont,
graphicx,
multirow,
booktabs,
amsmath,
amssymb,
hhline,
url}
\usepackage{marvosym}
\usepackage[ruled]{algorithm2e}
\usepackage[table,xcdraw]{xcolor}
\usepackage{tikz}
\usepackage{setspace}
\usepackage[edges]{forest}
\usetikzlibrary{trees,positioning,shapes,shadows,arrows.meta}
\usepackage{tabularx} 
\usepackage{makecell}
\usepackage[pagebackref=true,breaklinks=true,colorlinks,bookmarks=false,citecolor=blue,linkcolor=blue,urlcolor=blue]{hyperref}
\usepackage{ragged2e}

\definecolor{lightcyan}{rgb}{0.8, 1.0, 1.0}

\definecolor{hidden-draw}{RGB}{195,82,66}
\definecolor{hidden-blue}{RGB}{194,232,247}
\definecolor{hidden-orange}{RGB}{195,82,66}
\definecolor{hidden-yellow}{RGB}{242,244,193}
\definecolor{tree-level-1}{RGB}{195,82,66}
\definecolor{tree-level-2}{RGB}{195,82,66}
\definecolor{tree-level-3}{RGB}{195,82,66}
\definecolor{tree-leaf}{RGB}{195,82,66}

\definecolor{darkblue}{rgb}{0, 0, 0.6}
\definecolor{darkgreen}{rgb}{0, 0.7, 0}
\definecolor{darkred}{rgb}{0.8, 0, 0}
\definecolor{lightgreen}{rgb}{0.13, 0.55, 0.13}

\definecolor{ziren}{rgb}{255, 0, 0}
\definecolor{guo}{rgb}{0, 0, 0.6}

\let\oldcite\cite
\renewcommand{\cite}[1]{\textcolor{darkblue}{\oldcite{#1}}}

\newcommand*\colorcheck{%
  \expandafter\newcommand\csname greencheck\endcsname{\textcolor{darkgreen}{\ding{52}}}%
}
\newcommand*\colorcross{%
  \expandafter\newcommand\csname redcross\endcsname{\textcolor{darkred}{\ding{56}}}%
}
\colorcheck
\colorcross

\definecolor{firstcolor}{HTML}{BDE6CD}%{8EE574}
\definecolor{secondcolor}{HTML}{E2EEBC}%{FCFF91}
\definecolor{thirdcolor}{HTML}{FFF8C5}%{FFC5BF}

\newcommand{\fst}[1]{\cellcolor{firstcolor}\bfseries #1}
\newcommand{\snd}[1]{\cellcolor{secondcolor}#1}
\newcommand{\trd}[1]{\cellcolor{thirdcolor}#1}

\begin{document}
% \title{\textcolor{red}{Dynanic Scene Reconstruction: A Comprehensive Survey}}
\title{\textcolor{black}{Reconstructing the Dynamic World: \\ A Representation-Centric View of \\ 4D Scene Reconstruction}}

\author{
Ziren Gong, Guo Chen, Yongjia Li, Yihua Shao, Fabio Tosi,
Stefano Mattoccia, Matteo Poggi, Hao Tang\textsuperscript{\dag},
Fei Ma, Shuyan Li, Ziyang Yan\textsuperscript{\dag}, Nicu Sebe,
\IEEEmembership{Senior Member, IEEE},
Ling Shao, \IEEEmembership{Fellow, IEEE},\\
Jianfei Cai, \IEEEmembership{Fellow, IEEE},
Qi Tian, \IEEEmembership{Fellow, IEEE},
and Ming-Hsuan Yang, \IEEEmembership{Fellow, IEEE}
\IEEEcompsocitemizethanks{
\IEEEcompsocthanksitem Ziren Gong, Fabio Tosi, Stefano Mattoccia, and Matteo Poggi are with the Department of Computer Science and Engineering, University of Bologna, Italy. 

Guo Chen is with the Wangxuan Institute of Computer Technology, Peking University, China.

Yongjia Li and Yihua Shao are with the Department of Computing, The Hong Kong Polytechnic University, Hong Kong.

Hao Tang is with the School of Computer Science, Peking University, China.

Fei Ma and Qi Tian are with Guangdong Laboratory of Artificial Intelligence and Digital Economy (SZ), China. Qi Tian is also with Huawei.

% Fabio Remondino and Ziyang Yan are with the 3D Optical Metrology unit, Fondazione Bruno Kessler, Italy.

Shuyan Li is with the School of Electronics, Electrical Engineering and Computer Science, Queen's University Belfast, United Kingdom.

Ziyang Yan and Nicu Sebe are with the Department of Information Engineering and Computer Science, University of Trento, Italy.

Ling Shao is with the University of the Chinese Academy of Sciences, China.

Jianfei Cai is with the Faculty of IT, Monash University, Australia.

Ming-Hsuan Yang is with Google DeepMind and the University of California, Merced, United States.

\IEEEcompsocthanksitem Corresponding author: Ziyang Yan, Hao Tang.
\IEEEcompsocthanksitem E-mail:
yanziyang199634@gmail.com; bjdxtanghao@gmail.com}
\thanks{\dag~denotes corresponding authors.}
}

\IEEEtitleabstractindextext{
\justify
\begin{abstract}
4D scene reconstruction aims to recover the evolving geometry, appearance, and motion of dynamic environments from visual observations. Despite substantial progress in neural scene representations, reconstructing dynamic scenes remains challenging due to non-rigid motion, occlusions, temporal inconsistencies, and the trade-offs between reconstruction fidelity and computational efficiency. Recent advances in Neural Radiance Fields (NeRF) and 3D Gaussian Splatting (3DGS) have introduced diverse approaches to representing and reconstructing dynamic scenes, yet their relationships, underlying design choices, and evaluation protocols remain fragmented. In this paper, we present a unified perspective on 4D scene reconstruction, organizing existing methods around their scene representations, temporal modeling strategies, reconstruction pipelines, and optimization objectives. Through this framework, we examine how different design choices affect geometric fidelity, appearance consistency, motion representation, and computational efficiency. We further consolidate commonly used datasets and evaluation metrics, identify limitations in current experimental practices, and discuss open challenges in reconstructing complex, dynamic real-world environments. By connecting methodological developments with their underlying assumptions and evaluation evidence, this work provides a structured foundation for understanding existing approaches and identifying future research directions. An evolving collection of relevant papers and resources is available at \href{https://github.com/ZiyangYan/Awesome-4D-Scene-Reconstruction}{the project webpage}.

\end{abstract}

\begin{IEEEkeywords}
4D Scene Reconstruction, Neural Radiance Field, Gaussian Splatting, Methodological Evaluation
\end{IEEEkeywords}}

\maketitle

\input{sec/1_introduction}
\input{sec/2_background}
\input{sec/3_methodology_nerf}

\input{sec/4_methodology_gs}
\input{sec/5_dataset_and_benchmark}

\input{sec/6_4D_future}
\input{sec/7_conculsion}

\end{document}

%% file: sec/1_introduction.tex
\section{Introduction}
\label{sec:Introduction}
Scene reconstruction is a core problem in robotics and spatial AI, aiming to recover the three-dimensional structure and appearance of real-world environments from multi-view observations. It remains a fundamental challenge in computer vision and graphics, with applications spanning navigation, scene understanding, and novel view synthesis~\cite{zhang2025advances,he2025survey,Yan_2026_WACV,li2025near,yan2025evaluating}. Considerable efforts have been devoted to developing methods for dense, accurate, and high-fidelity reconstruction.

The field has evolved substantially over the past three decades. Early approaches were based on classical geometric pipelines, with Structure from Motion (SfM) and Multi-View Stereo (MVS) forming the foundation of three-dimensional reconstruction. SfM estimates camera poses and sparse scene geometry via feature matching and bundle adjustment, while MVS densifies these reconstructions using photometric consistency across calibrated views, producing detailed surface models of static scenes~\cite{schonberger2016structure}. Despite their effectiveness, these methods are limited by computational efficiency and their reliance on hand-crafted features and geometric assumptions.

With the development of Simultaneous Localization and Mapping (SLAM)~\cite{cheeseman1987stochastic}, scene reconstruction has progressed from offline batch processing to online frameworks that jointly perform camera tracking and environment mapping. However, traditional geometric pipelines remain sensitive to noise in incremental inputs, leading to cumulative trajectory drift and motion-induced artifacts that degrade reconstruction quality over time~\cite{nobre2017drift,yan2025evaluating}.

\begin{figure*}[tp]
    \centering
    \includegraphics[width=\linewidth]{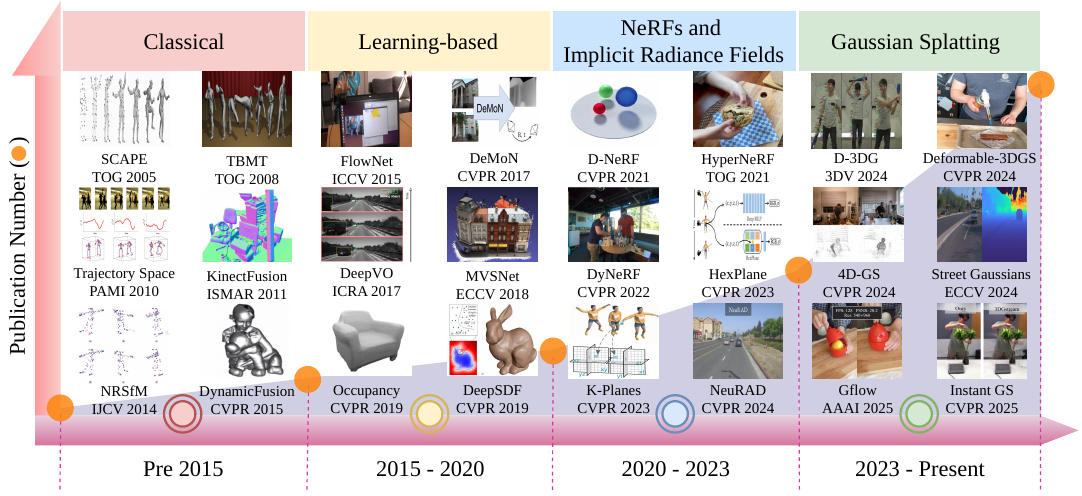}
     \vspace{-15pt}
\caption{\textbf{Trends in 4D reconstruction.} The increasing adoption of NeRF- and GS-based methods for dynamic scene modeling has led to a rapid growth in related publications in recent years.}
    \label{fig:timeline}
    \vspace{-10pt}
\end{figure*}

The field has undergone a paradigm shift with the advent of deep learning and neural representations. Neural Radiance Fields (NeRF)~\cite{mildenhall2021nerf} represent scenes as continuous implicit functions, enabling high-quality view synthesis. Subsequent extensions have improved training efficiency, rendering speed, and robustness~\cite{gao2022nerf,remondino2023critical,yan2023nerfbk}. More recently, 3D Gaussian Splatting (3DGS)~\cite{kerbl20233d} has emerged as an effective alternative, representing scenes with anisotropic 3D Gaussians and leveraging a differentiable tile-based rasterizer for real-time rendering while preserving fine details. These approaches depart from discrete geometric representations and instead learn continuous scene representations that capture complex geometry and appearance.

% Despite these advances, extending reconstruction to dynamic environments remains challenging. Key issues include non-rigid deformation and motion blur, which arise from fast motion and object dynamics; temporal consistency, which requires maintaining coherent surface correspondences over time; and appearance variation due to changing illumination and motion artifacts.
Despite advances in static scene modeling, extending reconstruction to dynamic environments introduces substantially greater complexity. High-fidelity 4D reconstruction must address the ambiguities of non-rigid motion, preserve long-range spatio-temporal coherence, and disentangle time-varying radiance from geometric deformation.

% The emergence of 4D dynamic scene reconstruction reflects the extension of 3D modeling to include the temporal dimension $t$. Existing approaches can be broadly categorized into two classes: implicit 4D representations~\cite{li2021dnerf, nr-nerf, NeRFPlayer}, which model scenes via canonical representations and time-dependent deformation fields, and explicit 4D Gaussian-based methods~\cite{yang2023gs4d, 4DRotorGS}, which incorporate temporal modeling into Gaussian primitives to capture scene evolution over space and time.

The rapid development of 4D dynamic reconstruction, driven by the transition from implicit neural representations to explicit Gaussian primitives, has led to an increasingly diverse and fragmented research landscape~\cite{yan20263d}. As illustrated in Fig.~\ref{fig:timeline}, the field is undergoing a shift in design choices, with evolving trade-offs among rendering speed, temporal consistency, and memory efficiency. This survey is motivated by the need to consolidate recent advances into a coherent framework and provide a structured reference for both established methods and emerging approaches such as 4D Gaussian Splatting (4DGS). 

%Recent years have witnessed rapid progress in this field, resulting in a fragmented literature with diverse methodologies developed largely in isolation. 
Existing surveys~\cite{gao2022nerf,xie2022neural,fei20243d,wu2024recent,bao20253d,chen2024survey,tosi2024nerfs} primarily focus on static neural rendering and often treat dynamic reconstruction as a secondary topic. As a result, a systematic review of 4D dynamic reconstruction using scene-specific optimization remains lacking.
In this survey, we focus on the two principal paradigms for scene-specific optimization in dynamic reconstruction: implicit 4D Neural Radiance Fields (4D NeRF) and explicit 4D Gaussian Splatting (4DGS). While classical Simultaneous Localization and Mapping (SLAM) and Structure from Motion (SfM) provide the geometric foundation for camera tracking and mapping, they are typically predicated on rigid-body assumptions or sparse representations. Emerging feed-forward, generalizable 4D reconstruction approaches enable rapid, offline-style inference across diverse scenes; however, they are fundamentally constrained by reliance on pre-trained dataset biases, often resulting in a \textit{fidelity ceiling} in out-of-distribution scenarios. 
We emphasize per-scene optimization approaches, which remain the gold standard for achieving high-fidelity reconstruction and temporal consistency. Both 4D NeRF and 4DGS mitigate the generalization gap of feed-forward models by anchoring reconstruction to scene-specific observations, rather than the statistical priors of a training set.
Furthermore, we analyze the performance of representative methods on benchmark datasets and discuss key open challenges for future research.

%% file: sec/2_background.tex
\section{Preliminaries}
\label{sec:Background}
\subsection{History of Dynamic Scene Reconstruction}
\label{subsec:Dynamic Scene Reconstruction}
As shown in Fig.~\ref{fig:timeline}, dynamic scene reconstruction has evolved over the past two decades from geometry-based pipelines to learning-based methods and, more recently, to implicit neural representations. The introduction of NeRF established 4D reconstruction as a central research direction, further accelerated by 3DGS. This has led to rapid growth in recent work. This section reviews this progression and highlights key developments underlying modern radiance field and Gaussian-based approaches.

\noindent \textbf{Classical Approaches: From Geometry to Volumetric Fusion (Pre~2015).}
Early work was dominated by geometry-based pipelines. Multi-view stereo (MVS) and structure-from-motion (SfM) were extended to dynamic settings, leading to non-rigid SfM (NRSfM)~\cite{NRSfM} and template-based mesh tracking~\cite{de2008performance}. These methods estimate per-frame geometry and track deformations relative to a canonical model. In parallel, depth sensors enabled volumetric fusion methods, such as KinectFusion~\cite{newcombe2011kinectfusion} and DynamicFusion~\cite{newcombe2015dynamicfusion}, which integrate depth maps into a canonical volume with non-rigid alignment.

These approaches have several limitations. Explicit geometry restricts modeling of complex deformations and topology changes. Per-frame optimization limits scalability for long sequences and high-resolution scenes. Appearance modeling is also limited, typically relying on texture maps and lacking view-dependent effects.

\noindent \textbf{Transition to Learning-based Models (2015--2020).}
With the rise of deep learning, reconstruction methods began incorporating learned priors. Early works~\cite{yao2018mvsnet,ummenhofer2017demon,dosovitskiy2015flownet,ilg2017flownet,vijayanarasimhan2017sfm,wang2017deepvo} improved components such as depth, scene flow, and pose estimation, while retaining classical pipelines. A key shift was the adoption of implicit representations, including occupancy networks~\cite{mescheder2019occupancy} and neural SDFs~\cite{park2019deepsdf}, which model scenes as continuous fields. These representations are more expressive and compact, and enable dynamic extensions.

\noindent \textbf{Dynamic NeRFs and Implicit Radiance Fields (2020--2023).}
Neural Radiance Fields (NeRF)~\cite{mildenhall2021nerf} enable high-quality novel view synthesis by modeling scenes as continuous radiance fields. Dynamic variants, such as D-NeRF~\cite{li2021dnerf}, Nerfies~\cite{park2021nerfies}, and HyperNeRF~\cite{park2021hypernerf}, represent scenes using canonical fields with deformation models. These methods capture complex non-rigid motion.
However, dynamic NeRFs are computationally expensive. Training often requires long runtimes, and inference is slow due to volumetric rendering. Temporal consistency remains difficult over long sequences. Later methods, including TiNeuVox~\cite{fang2022fast} and NSFF~\cite{nsff}, introduce explicit structures or motion priors to improve efficiency, but scalability remains limited.

\noindent \textbf{Dynamic 3D Gaussian Splatting (2023--Present).}
Gaussian Splatting (GS)~\cite{kerbl20233d} represents scenes as sets of anisotropic 3D Gaussians with learnable parameters, including position, scale, opacity, rotation, and appearance. Dynamic extensions incorporate temporal modeling through deformation or per-frame transformations, leading to 4D Gaussian Splatting (4DGS).
GS enables real-time rendering via rasterization. Its explicit structure facilitates handling occlusions and non-rigid motion. Each Gaussian jointly encodes geometry and appearance, yielding a compact representation.
However, dynamic GS methods often require many primitives, leading to high memory usage~\cite{hu20254dgc, yuan20251000+}. Improving efficiency and scalability remains an open problem.

\begin{figure}[t]
\centering
\includegraphics[width=0.6\linewidth]{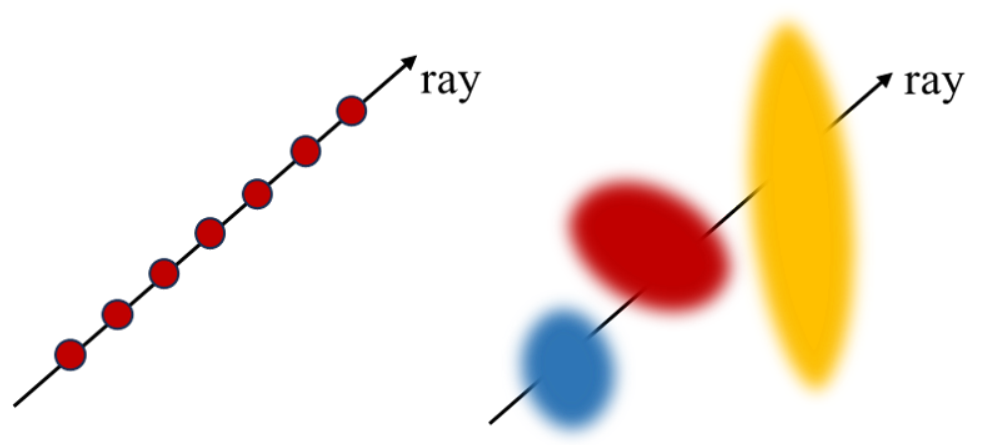}
\vspace{-10pt}
\caption{\textbf{Comparison between NeRF and 3DGS}. NeRF (left) evaluates an MLP along each ray, whereas 3DGS (right) renders by rasterizing and blending Gaussians.}
\label{fig:nerf_3dgs_diff}
    \vspace{-10pt}
\end{figure}

%HERE
\subsection{Representing Radiance Fields}

Recent advances in radiance field representations have improved 4D scene reconstruction, enabling high-fidelity modeling of geometry and appearance over time. Both NeRF and 3DGS represent scenes as radiance fields but differ in formulation. NeRF models the scene implicitly using a neural network that maps 3D coordinates and viewing directions to color and density. In contrast, 3DGS represents the scene explicitly with anisotropic 3D Gaussians optimized for efficient rendering. We briefly review these approaches and summarize their differences in Fig.~\ref{fig:nerf_3dgs_diff}.

\noindent \textbf{Neural Radiance Fields.} NeRF~\cite{mildenhall2021nerf} models a 3D scene as a continuous function that maps a 3D point and viewing direction to color and density:
\begin{equation}
F_{\theta}(\mathbf{x}, \mathbf{d}) \rightarrow (\mathbf{c}, \sigma).
\end{equation}

\noindent Novel views are synthesized via differentiable volume rendering, where pixel color is computed by integrating radiance along a camera ray:
\begin{equation}
\hat{C}(\mathbf{r})=\sum_{i=1}^{N} T_i \left(1-e^{-\sigma_i \delta_i}\right)\mathbf{c}_i, \quad 
T_i=\exp\left(-\sum_{j=1}^{i-1}\sigma_j \delta_j\right).
\end{equation}

\noindent To capture high-frequency details, NeRF applies positional encoding to the input coordinates, enabling the MLP to represent complex geometry and appearance.

\noindent \textbf{3D Gaussian Splatting.} 3DGS~\cite{kerbl20233d} represents a scene explicitly as a set of anisotropic 3D Gaussians. Each Gaussian encodes position, density, and appearance, analogous to volumetric elements in NeRF but in an explicit form. Initialized from a sparse SfM point cloud, each Gaussian is parameterized by its mean~$\mu$ and covariance~$\Sigma$:
\begin{equation}
G(x) = \exp\!\left(-\tfrac{1}{2}(x-\mu)^\top \Sigma^{-1} (x-\mu)\right), \quad \Sigma = R S S^\top R^\top.
\end{equation}

\noindent For rendering, Gaussians are projected to the image plane and composited via alpha blending:
\begin{equation}
C = \sum_{i} c_i \alpha_i \prod_{j<i}(1-\alpha_j).
\end{equation}

\noindent Each Gaussian defines a continuous density in space and contributes to pixel color along a camera ray, analogous to volumetric rendering in NeRF.

\subsection{Comparison with Existing Surveys}

Recent advances in radiance field representations have driven the development of 4D reconstruction methods based on NeRF and 3DGS, with applications across diverse domains. Table~\ref{tab:survey_comparison} compares recent surveys in this area. Our survey provides broader coverage by including both NeRF- and GS-based methods, summarizing commonly used datasets and evaluation protocols, and incorporating a quantitative analysis for performance comparison. 

Among existing works, Zhu et al.~\cite{zhu2025dynamic} is most closely related, but covers fewer methods (52 vs.~102) and datasets (10 vs.~22), and omits several scene types, such as autonomous driving. In addition, it does not provide a unified taxonomy or a comprehensive evaluation. This survey aims to offer a structured and comprehensive reference for 4D scene reconstruction.

\begin{table*}[tp]
\centering
\caption{\textbf{Comparison of existing surveys on 4D scene reconstruction.}}
\resizebox{\textwidth}{!}{%
\rowcolors{2}{lightcyan}{white}
\begin{tabular}{lcccccc}
    \hline
    \textbf{Survey}  & \textbf{4D Scene Types}  &  
    \textbf{\makecell{Evaluation\\Coverage}} &  \textbf{Datasets} & \textbf{Methods} & \textbf{Taxonomy} \\ \hline

    Fan et al.~\cite{fan2025advances}  
    & \makecell{Human and animal motion}  
    & -- &  -- & 90 & -- \\ 

    Zhu et al.~\cite{zhu2025dynamic} 
    & General   & NVS, Efficiency & 10 & 52 & \checkmark \\ 

    He et al.~\cite{he2024neural}  
    & Autonomous driving 
    & -- & -- & 36 & -- \\ 

    Cao et al.~\cite{cao2025reconstructing}  
    & General 
    & -- & -- & 111 & -- \\ 

    Zhao et al.~\cite{zhao2025advances}  
    & \makecell{Object, human, and animal motion}   
    & -- & 21 & 62 & -- \\ 

    Ours 
    & General
    & NVS, Geometry, Efficiency 
    & 22 & 102 & \checkmark  \\ 
    \hline
\end{tabular}}
\label{tab:survey_comparison}
\end{table*}

%% file: sec/3_methodology_nerf.tex
% \section{Dynamic Scene Reconstruction with NeRF}

%HERE
\section{DYNAMIC NEURAL RADIANCE FIELDS}
\label{sec:NeRF}

\begin{figure*}[t]
\centering
\includegraphics[width=\textwidth]{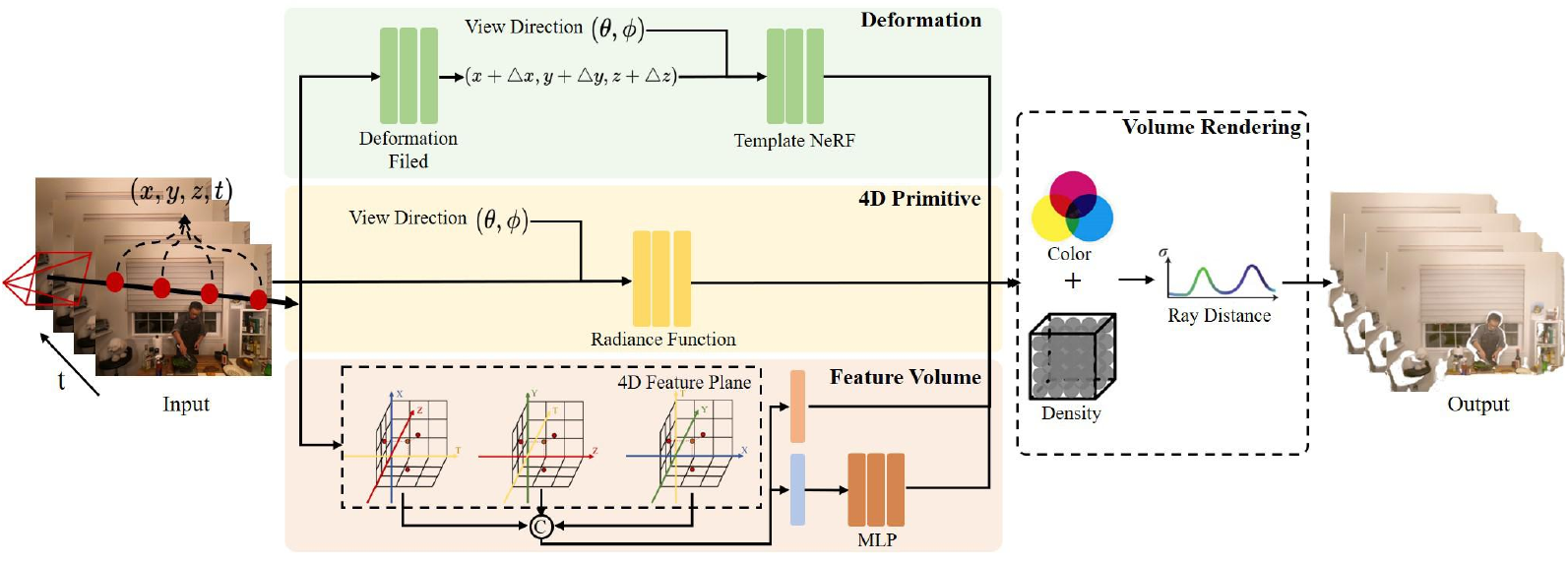}
\vspace{-10pt}
\caption{General pipeline of \textbf{NeRF-based} 4D scene reconstruction methods. The pipeline illustrates representative strategies, including deformation-based, 4D primitive-based, and 4D feature volume-based frameworks. Temporal prior-based methods are not included due to their diversity.}
\label{nerf_pipeline}
    \vspace{-10pt}
\end{figure*}

\subsection{Preliminaries of 4D NeRF}

% To extend static NeRF to dynamic scene, modeling temporal dynamics requires fundamental modifications to handle the additional time dimension. The incorporation of time into NeRF architectures has led to the development of four primary methodological categories, each with distinct approaches to representing and processing temporal information within the neural radiance field framework.
To extend static NeRF to dynamic scenes, temporal dynamics are incorporated by introducing a time dimension into the radiance field formulation. As shown in Fig.~\ref{nerf_pipeline}, existing approaches adopt different strategies to encode temporal information within neural radiance fields. Based on the taxonomy in Table~\ref{tab:4dnerf}, these methods can be categorized into four main classes.

\noindent \textbf{Deformation-Field-Based Methods} extend static NeRF by maintaining a canonical 3D representation and learning time-dependent spatial transformations. This approach is based on the observation that many dynamic scenes can be modeled as deformations of a canonical state, where geometric changes are captured by learnable displacement functions. The formulation decomposes the 4D problem into a canonical NeRF and a deformation network. The canonical NeRF $F_{\theta}$ models a reference frame, while the deformation network $D_{\phi}$ maps spatial coordinates at time $t$ to the canonical space:
\begin{equation}
\begin{aligned}
\mathbf{x}' &= D_{\phi}(\mathbf{x}, t), \\
F_{\theta}: (\mathbf{x}', \mathbf{d}) &\rightarrow (\mathbf{c}, \sigma),
\end{aligned}
\end{equation}
where $\mathbf{x}'$ denotes the canonical coordinate and $t$ is the temporal variable. This formulation leverages static NeRF priors while introducing a compact mechanism to model temporal variations.

\noindent \textbf{Implicit 4D Primitive-Based Methods} model time as an additional input alongside spatial coordinates and viewing direction. This formulation extends NeRF to a unified 4D representation without explicit decomposition into canonical and deformation components. The radiance field directly maps space-time coordinates to color and density:
\begin{equation}
F_{\theta}: (\mathbf{x}, \mathbf{d}, t) \rightarrow (\mathbf{c}, \sigma),
\end{equation}
where the temporal variable $t$ is encoded jointly with spatial inputs, enabling the model to capture temporal variations. This formulation offers high flexibility for modeling complex dynamics but increases computational cost and requires careful temporal sampling.

\noindent \textbf{4D Feature-Volume-Based Methods} improve efficiency by decomposing the space-time domain into structured representations. The key idea is to approximate the 4D volume using lower-dimensional components, reducing memory and computation while preserving expressiveness. A common approach factorizes the 4D space using plane-based representations, such as tri-plane extensions:
\begin{gather}
\begin{aligned}
    \mathbf{f}(\mathbf{x}, t) &= \mathbf{f}_{xy}(x, y) \odot \mathbf{f}_{xz}(x, z) \odot \mathbf{f}_{yz}(y, z) \\
    &\quad \odot \mathbf{f}_{xt}(x, t) \odot \mathbf{f}_{yt}(y, t) \odot \mathbf{f}_{zt}(z, t),
\end{aligned} \\
F_{\theta}: \mathbf{f}(\mathbf{x}, t) \rightarrow (\mathbf{c}, \sigma),
\end{gather}
where $\odot$ denotes element-wise multiplication and $\mathbf{f}_{xy}, \mathbf{f}_{xt}, \mathbf{f}_{yt}$ denote spatial and spatiotemporal feature planes. Alternative approaches use tensor decomposition (e.g., CP or Tucker) to represent the 4D volume with low-rank components, enabling efficient storage and rendering while maintaining temporal coherence.

\noindent \textbf{Temporal-Prior-Based Methods} incorporate external temporal cues as auxiliary supervision, rather than explicitly parameterizing time within the radiance field. The key idea is to enforce temporal coherence through additional constraints derived from complementary signals, without modifying the underlying NeRF formulation. These methods augment standard training with temporal consistency losses and motion priors:
\begin{equation}
\begin{aligned}
F_{\theta}: (\mathbf{x}, \mathbf{d}, t) &\rightarrow (\mathbf{c}, \sigma), \\
\mathcal{L} &= \mathcal{L}_{\text{recon}} + \lambda_1 \mathcal{L}_{\text{flow}}(t) + \lambda_2 \mathcal{L}_{\text{temp}}(t),
\end{aligned}
\end{equation}
where $\mathcal{L}_{\text{flow}}$ encodes motion cues (e.g., optical or scene flow) to enforce geometric consistency, and $\mathcal{L}_{\text{temp}}$ enforces temporal smoothness. These approaches improve coherence and can be combined with other designs.

\begin{table*}[tp]
\caption{\textbf{Overview of NeRF-based 4D dynamic scene reconstruction methods}. Methods are categorized into four types. For each method, we summarize its scene representation, key components, and additional priors.}
\label{tab:4dnerf}
\centering
\resizebox{\textwidth}{!}{%
%\rowcolors{2}{gray!15}{white}
\rowcolors{2}{lightcyan}{white}
\begin{tabular}{lcccccccccc}
\hline
\multicolumn{1}{l}{Method} &
  \multicolumn{1}{c}{Venue} &
  \multicolumn{1}{c}{Inputs} &
  \multicolumn{1}{c}{Scenario} &
  \multicolumn{1}{c}{Target Domain} &
  4D-style &
  Scene Encoding &
  \multicolumn{1}{c}{Flow} &
  Normal &
  Segment. &
  Extra Prior \\ \hline
% \multicolumn{11}{c}{NeRF-Style} \\ \hline
D-NeRF~\cite{li2021dnerf}          & CVPR2021            & RGB & Indoor    & Entity-Centric & Deformation Fields  & MLP                 &   &   &   &                               \\
NR-NeRF~\cite{nr-nerf}          & CVPR2021            & RGB & In-the-Wild   & Scene-Centric & Deformation Fields  & MLP                  &   &   &   &                               \\
STaR~\cite{star}             & CVPR2021            & RGB & Indoor    & Scene-Centric & Deformation Fields  & MLP                  &   &   &   &                               \\
Nerfies \cite{park2021nerfies} & ICCV2021 & RGB & Indoor & Entity-Centric & Deformation Fields & MLP & & \multicolumn{1}{l}{} & \multicolumn{1}{l}{} & \multicolumn{1}{l}{} \\
HyperNeRF \cite{park2021hypernerf} & TOG2021 & RGB & Indoor & Entity-Centric & Deformation Fields & MLP &  & \multicolumn{1}{l}{} & \multicolumn{1}{l}{} & \multicolumn{1}{l}{} \\
NDR~\cite{NDR}               & NIPS2022            &RGBD & Indoor    & Entity-Centric  & Deformation Fields  & MLP                  &   &   &   &                               \\
DeVRF~\cite{liu2022devrf}             & RGBNIPS2022            & RGB & Indoor    & Entity-Centric & Deformation Fields  & Voxel Grid + MLP    & \checkmark &   &   & RAFT                         \\
TiNeuVox~\cite{fang2022fast}          & ACM SIG. 2022            & RGB & Indoor    & Entity-Centric & Deformation Fields  & MLP           &   &  &   &   \\
NeRF-DS~\cite{NeRF-DS}          & CVPR2023            & RGB & Indoor    & Entity-Centric & Deformation Fields  & MLP                  &   & \checkmark &   &                               \\
RoDynRF~\cite{RoDynRF}          & CVPR2023            & RGB & In-the-Wild   & Scene-Centric & Deformation Fields  & Voxel Grid + MLP    & \checkmark &   &   & RAFT                         \\
Total-Recon~\cite{gao2023totalrecon}      & ICCV2023            & RGBD & Indoor    & Scene-Centric & Deformation Fields  & MLP                 & \checkmark &   &   & VCN                          \\
DyBluRF~\cite{sun2024dyblurf}          & CVPR2024            & RGB & In-the-Wild   & Scene-Centric & Deformation Fields  & MLP                  & \checkmark &   &   & RAFT                         \\
NSFF~\cite{nsff}             & CVPR2021            & RGB & In-the-Wild   & Scene-Centric & 4D Primitive         & MLP                & \checkmark &   &   & RAFT                         \\
Video-NeRF~\cite{video-nerf}       & CVPR2021            & RGBD & In-the-Wild     & Entity-Centric  & 4D Primitive         & MLP                 &   &   &   &                               \\
DynNeRF~\cite{dynnerf}          & ICCV2021            & RGB & In-the-Wild     & Scene-Centric & 4D Primitive         & MLP                  & \checkmark &   &   & RAFT                         \\
NeRFlow~\cite{nerflow}          & ICCV2021            & RGB & Indoor    & Entity-Centric & 4D Primitive         & MLP                  & \checkmark &   &   & Farneback G.                  \\
DyNeRF \cite{li2022neural} & CVPR2022 & RGB & Indoor & Scene-Centric & 4D Primitive & MLP & & \multicolumn{1}{l}{} & \multicolumn{1}{l}{} & \multicolumn{1}{l}{} \\
MonoNeRF~\cite{tian2023mononerf}         & ICCV2023            & RGB & In-the-Wild   & Scene-Centric & 4D Primitive         & MLP                 & \checkmark &   &   & RAFT                         \\
Sync-NeRF~\cite{sync-nerf}        & AAAI2024            & RGB & Indoor    & Scene-Centric & 4D Primitive         & MLP                  &   &   &   &                               \\
4DNDF~\cite{4d-ndf}             & CVPR2024            & LiDAR & Auto. Driving   & Scene-Centric & 4D Primitive         & Hash Grid + MLP    &   &   & \checkmark &                               \\
Ml-nsg~\cite{ml-nsg}            & CVPR2024            & RGB & Auto. Driving   & Scene-Centric & 4D Primitive         & Hash Grid + MLP    &   &   &   &                               \\
DecouplingNeRF~\cite{decouplingnerf}   & TVCG2024            & RGB & In-the-Wild    & Scene-Centric& 4D Primitive         & MLP                  & \checkmark &   &   & NSFF                         \\
DetNeRF~\cite{detnerf}          & AAAI2025            & RGB & In-the-Wild   & Scene-Centric & 4D Primitive         & MLP           & \checkmark &   &   & RAFT                         \\
HexPlane~\cite{cao2023hexplane}         & CVPR2023            & RGB & Indoor    & Entity-Centric & 4D Feature Volumes  & Feat. Plane + MLP   &   &   &   &                               \\
K-Planes~\cite{K-Planes}         & CVPR2023            & RGB & Indoor    & Entity-Centric & 4D Feature Volumes  & Feat. Plane + MLP  &   &   &   &                               \\
SUDS~\cite{suds}             & CVPR2023            & RGBD & Auto. Driving   & Scene-Centric & 4D Feature Volumes  & Hash Grid + MLP    & \checkmark &   &   & RAFT \& DINO                 \\
TIDNeRF~\cite{TID-NeRF}          & CVPR2023            & RGB & Indoor    & Multi.-Centric & 4D Feature Volumes  & Hash Grid + MLP  &   &   &   &                               \\
HyperReel~\cite{attal2023hyperreel}        & CVPR2023            & RGB & Indoor    & Entity-Centric & 4D Feature Volumes  & Feat. Plane + MLP  &   &   &   &                               \\
MixVoxels~\cite{MixVoxels}        & ICCV2023            & RGB & In. \& Wild   & Scene-Centric & 4D Feature Volumes  & Voxel Grid + MLP    &   &   &   &                               \\
MSTH~\cite{MSTH}             & NIPS2023            & RGB & Indoor    & Scene-Centric & 4D Feature Volumes  & Hash Grid + MLP    &   &   &   & Kendall and Gal              \\
NVFi~\cite{nvfi}             & NIPS2023            & RGB & In. \& wild   & Multi.-Centric & 4D Feature Volumes  & Feat. Plane + MLP  &   &   &   & HexPlane                     \\
NeRFPlayer~\cite{NeRFPlayer}       & TVCG2023            & RGB & Indoor    & Scene-Centric & 4D Feature Volumes  & Hybrid + MLP   &   &   &   &                               \\
BLiRF~\cite{blirf}             & AAAI2024            & RGB & Indoor    & Multi.-Centric  & 4D Feature Volumes  & MLP       &   &   &   &                               \\
Ced-NeRF~\cite{Ced-NeRF}         & AAAI2024            & RGB & Indoor    & Multi.-Centric & 4D Feature Volumes  & Hash Grid + MLP   &   &   &   &                               \\
LiDAR4D~\cite{lidar4d}          & CVPR2024            & LiDAR & Auto. Driving   & Scene-Centric & 4D Feature Volumes  & Hybrid + MLP   & \checkmark &   &   & flow MLP                     \\
DaReNeRF~\cite{DaReNeRF}         & CVPR2024            & RGB & Indoor    & Scene-Centric & 4D Feature Volumes & Feat. Plane + MLP  &   &   &   &                        \\
Gear-NeRF~\cite{Gear-NeRF}        & CVPR2024            & RGB & Indoor    & Scene-Centric & 4D Feature Volumes & Feat. Plane + MLP   &   &   & \checkmark & SAM                          \\
NeuRAD~\cite{NeuRAD}            & CVPR2024            & RGBD & Auto. Driving   & Scene-Centric & 4D Feature Volumes & Hash Grid + MLP    &   &   &   &                               \\
S-DyRF~\cite{S-DyRF}            & CVPR2024            & RGB & Indoor    & Multi.-Centric & 4D Feature Volumes  & Feat. Plane + MLP  &   &   &   & HexPlane                     \\
RoDUS~\cite{RoDUS}             & ECCV2024            & RGB & Auto. Driving   & Scene-Centric & 4D Feature Volumes  & Hash Grid + MLP    & \checkmark &   & \checkmark &                               \\
EmerNeRF~\cite{EmerNeRF}         & ICLR2024            & RGBD & Auro. Driving   & Scene-Centric & 4D Feature Volumes  & Hash Grid + MLP   & \checkmark &   &   & DINOv2                       \\
SLS4D~\cite{SLS4D}             & TVCG2024            & RGB & Indoor    & Entity-Centric & 4D Feature Volumes  & Hybrid + MLP    &   &   &   &                               \\
StreamRF~\cite{StreamRF}         & NIPS2022            & RGB & Indoor    & Scene-Centric & Temporal Prior      & Voxel Grid + MLP   &   &   &   &                               \\
OTNeRF~\cite{OT-NeRF}           & ICLR2024            & RGB & Indoor    & Entity-Centric & Temporal Prior      & MLP        &   &   &   &                               \\
STGC-NeRF~\cite{STGC-NeRF}        & AAAI2025            & LiDAR & Auro. Driving   & Scene-Centric & Temporal Prior      & Hier. Repre. + MLP  & \checkmark & \checkmark &   & GMSF                         \\ \hline
\end{tabular}%
}
\end{table*}

\subsection{Deformation-Field-Based Methods}

These methods extend static NeRF by maintaining a canonical 3D representation and learning deformation functions $D(\mathbf{x}, t) \rightarrow \mathbf{x}'$ to model temporal variations.
To address non-rigid reconstruction without explicit geometry, early methods employ deformation MLPs to map spatio-temporal coordinates into a canonical space, e.g., \textbf{D-NeRF}~\cite{li2021dnerf}.  On the other hand, \textbf{NR-NeRF}~\cite{nr-nerf} introduces rigidity regularization to improve temporal correspondence. However, these methods remain less effective in the presence of significant topological changes or large displacements due to the limitations of a single continuous deformation field.
% \textbf{D-NeRF}~\cite{li2021dnerf} introduces a deformation MLP that maps spatiotemporal points to a canonical space, enabling reconstruction of both rigid and non-rigid motion from monocular input and supporting mesh extraction. Building on this idea, \textbf{NR-NeRF}~\cite{nr-nerf} incorporates a rigidity network and regularization to stabilize static regions and produce dense correspondences.

For complex scenes with articulated structures or multiple interacting objects, global deformation fields are often insufficient to capture localized motions. \textbf{STAR}~\cite{star}, \textbf{Total-Recon}~\cite{gao2023totalrecon}, and \textbf{NDR}~\cite{NDR} address this limitation by factorizing scenes into motion-aware canonical subspaces. With trajectory-based constraints, these methods jointly optimize geometry, camera poses, and non-rigid transformations. However, the optimization process remains computationally expensive and sensitive to the quality of the initial trajectory or pose estimates.
% For articulated and multi-object scenes, hierarchical motion modeling is required. \textbf{STAR}~\cite{star}, \textbf{Total-Recon}~\cite{gao2023totalrecon}, and \textbf{NDR}~\cite{NDR} impose trajectory-based constraints and factorize scenes into motion-aware canonical spaces, enabling joint optimization over RGB-D sequences and camera poses.

Reconstructing high-fidelity dynamic radiance fields from sparse observations or uncalibrated cameras remains challenging. To improve stability, \textbf{DeVRF}~\cite{liu2022devrf} and \textbf{RoDynRF}~\cite{RoDynRF} employ voxel-based representations for efficient canonicalization. These methods further incorporate auxiliary priors, such as monocular depth, disparity, and reprojection constraints, to jointly estimate camera motion and dynamic scene evolution. However, their performance depends heavily on the quality of the external priors, and inaccurate disparity estimates can introduce artifacts into the 4D representation.
% To improve robustness under sparse observations, \textbf{DeVRF}~\cite{liu2022devrf} and \textbf{RoDynRF}~\cite{RoDynRF} jointly estimate camera parameters and dynamic radiance fields using voxel representations, deformation-based canonicalization, and auxiliary priors such as reprojection, disparity, and monocular depth.

Typical deformation models often struggle with view-dependent specularities and motion blur, leading to entangled geometry and appearance. To alleviate this issue, \textbf{DyBluRF}~\cite{sun2024dyblurf} and \textbf{NeRF-DS}~\cite{NeRF-DS} incorporate surface-normal conditioning and mask-guided deformation to separate transient appearance effects from scene geometry. However, modeling complex radiance variations remains challenging in regions with rapid motion or strong reflections, often resulting in blurred textures or residual artifacts.

\subsection{Implicit 4D Primitive-Based Methods}

These methods extend NeRF by directly modeling dynamics through an explicit temporal dimension, where the radiance field is defined as $F(\mathbf{x}, t, \mathbf{d}) \rightarrow (\sigma, \mathbf{c})$. This formulation avoids canonical decomposition and provides a unified space-time representation.
To improve temporal coherence in dynamic scenes, several methods incorporate explicit motion fields into volumetric representations. \textbf{NSFF}~\cite{nsff} introduced neural scene flow fields to jointly optimize geometry, radiance, and dense 3D motion, while \textbf{NeRFlow}~\cite{nerflow} coupled radiance fields with continuous flow for consistent monocular view synthesis. However, these methods remain sensitive to flow estimation errors, often producing artifacts or blurred geometry under rapid motion.
% \textbf{NSFF}~\cite{nsff} introduces neural scene flow fields to jointly model geometry, appearance, and dense 3D motion, enabling temporally consistent reconstruction and view synthesis across space and time. Similarly, \textbf{NeRFlow}~\cite{nerflow} couples a radiance field with a flow field, allowing consistent modeling of appearance, density, and motion from monocular video.

To address the under-constrained nature of monocular reconstruction, several works adopt static--dynamic decomposition. \textbf{DynNeRF}~\cite{dynnerf} and \textbf{Video-NeRF}~\cite{video-nerf} incorporate monocular depth priors to regularize geometry and appearance, enabling stable free-viewpoint rendering of dynamic content. \textbf{DetNeRF}~\cite{detnerf} further extends this by employing occlusion-aware modeling to explicitly separate static backgrounds from moving components. The efficacy of these methods is strictly bounded by the quality of external priors

For long sequences and complex scenes with multiple moving agents, \textbf{DecouplingNeRF}~\cite{decouplingnerf} and \textbf{ML-NSG}~\cite{ml-nsg} use hierarchical neural scene graphs to decompose scenes into object-centric components for scalable reconstruction. However, the hierarchical design introduces additional complexity, and these methods may struggle with objects exhibiting unpredictable motion or frequent occlusions.
% In order to reduce reliance on external priors, \textbf{DecouplingNeRF}~\cite{decouplingnerf} and \textbf{ML-NSG}~\cite{ml-nsg} organize scenes using hierarchical neural scene graphs, enabling scalable reconstruction for long sequences and complex environments.

Recent advances increasingly focus on modeling motion dynamics and incorporating non-RGB sensors. \textbf{Sync-NeRF}~\cite{sync-nerf} and \textbf{MonoNeRF}~\cite{tian2023mononerf} learn implicit velocity fields and feature correspondences to improve temporal alignment and robustness. Beyond RGB inputs, \textbf{4D-NDF}~\cite{4d-ndf} models LiDAR sequences using time-dependent signed distance functions (SDFs) to jointly reconstruct static structures and dynamic objects. However, velocity-based methods remain sensitive to temporal aliasing, while multimodal approaches are affected by modality-specific noise.

\subsection{4D Feature-Volume-Based Methods}

To improve efficiency, these methods replace implicit neural fields with structured representations that factorize the 4D space-time domain into lower-dimensional components. Common strategies include tensor decomposition, planar factorization, and multi-resolution grids, enabling faster training and rendering with reduced memory.

%HERE
To reduce the computational cost of coordinate-based MLPs, several methods decompose 4D space into lower-dimensional representations. \textbf{HexPlane}~\cite{cao2023hexplane} and \textbf{K-Planes}~\cite{K-Planes} project 4D volumes onto orthogonal 2D planes, while \textbf{HyperReel}~\cite{attal2023hyperreel} and \textbf{MixVoxels}~\cite{MixVoxels} combine voxel grids with ray-conditioned sampling and deformation modeling. These factorizations enable efficient high-quality rendering through structured memory layouts, but often involve a trade-off between memory usage and spatial resolution.
%\textbf{HexPlane}~\cite{cao2023hexplane}, \textbf{K-Planes}~\cite{K-Planes}, \textbf{HyperReel}~\cite{attal2023hyperreel}, and \textbf{MixVoxels}~\cite{MixVoxels} adopt plane- or voxel-based factorizations, combining feature planes or grids with ray-conditioned sampling and deformation modeling to achieve efficient and high-quality rendering.

To ensure smooth transitions between temporal snapshots, several methods incorporate temporal interpolation and frequency-aware modeling. \textbf{TID-NeRF}~\cite{TID-NeRF} integrates temporal modeling into explicit representations, while \textbf{BLiRF}~\cite{blirf} models radiance fields as band-limited signals using neural trajectory bases and low-rank spatial decomposition. \textbf{NeRFPlayer}~\cite{NeRFPlayer} decomposes scenes into static and dynamic components with sliding-window temporal encoding. \textbf{Ced-NeRF}~\cite{Ced-NeRF} improves generalization with hybrid grid-based representations, and \textbf{DaReNeRF}~\cite{DaReNeRF} encodes temporal information using direction-aware wavelet representations. However, high-frequency motion and abrupt topological changes are often blurred by the underlying spectral constraints or interpolation schemes.

For large-scale scenes, hash-based encodings improve scalability. \textbf{SUDS}~\cite{suds} and \textbf{MSTH}~\cite{MSTH} utilize multiresolution hash grids with uncertainty-aware masking to handle complex urban dynamics, while \textbf{NeuRAD}~\cite{NeuRAD} and \textbf{RoDUS}~\cite{RoDUS} incorporate rolling shutter compensation and semantic signals for robust performance in autonomous driving scenarios. 
\textbf{EmerNeRF}~\cite{EmerNeRF} and \textbf{LiDAR4D}~\cite{lidar4d} further extend these ideas to multimodal settings by using learned flow and temporal feature slots to aggregate information across sparse frames. However, hash-based representations may suffer from feature collisions in complex scenes, potentially introducing aliasing artifacts or geometric noise.
% \textbf{EmerNeRF}~\cite{EmerNeRF} adopts self-supervised static--dynamic decomposition with learned flow for multi-frame aggregation. 
% For specialized modalities and efficiency constraints, tailored designs are adopted. \textbf{LiDAR4D}~\cite{lidar4d} and \textbf{SLS4D}~\cite{SLS4D} exploit sparsity with temporal feature slots and hybrid spatial encodings to achieve compact representations. 

Hybrid representations support specialized tasks by incorporating domain-specific inductive biases. \textbf{NVFi}~\cite{nvfi} introduces physics-informed velocity fields for motion transfer and future-state prediction, while \textbf{Gear-NeRF}~\cite{Gear-NeRF} leverages semantic priors for object-level tracking and motion-aware sampling. Methods such as \textbf{S-DyRF}~\cite{S-DyRF} further enable temporally consistent stylization in dynamic scenes. However, the reliance on specialized priors can limit generalization across diverse scenarios.
\subsection{Temporal-Prior-Based Methods}
While previous approaches explicitly parameterize time within the radiance field, practical reconstruction often benefits from external temporal cues used as constraints or guidance. These methods leverage signals such as optical flow, physical constraints, or sequential modeling to regularize the under-constrained dynamic reconstruction problem.

\noindent To support real-time and streaming applications, several frameworks replace global optimization with temporally incremental updates. \textbf{StreamRF}~\cite{StreamRF}, for example, employs a grid-based representation with sequential updates, modeling temporal evolution through incremental changes to a base model and enabling efficient streaming via difference-based compression. However, these methods are prone to error accumulation, where small update misalignments propagate over long sequences.

For scenes with complex stochastic motion, \textbf{OTNeRF}~\cite{OT-NeRF} enforces temporal consistency by modeling scene dynamics as low-frequency shifts in pixel distributions. However, such statistical regularization often suppresses high-frequency geometric details.
In scenarios with sparse observations or severe occlusions, incorporating physics-based or geometric priors from non-RGB sensors becomes important. \textbf{STGC-NeRF}~\cite{STGC-NeRF} introduces spatio-temporal geometric constraints for LiDAR-based reconstruction, using scene flow to establish cross-frame correspondences and improve robustness under sparse inputs. However, the effectiveness of these constraints depends heavily on motion estimation quality, and large inter-frame displacements can introduce temporal aliasing in high-speed scenes.

% \textbf{StreamRF}~\cite{StreamRF} adopts a grid-based representation with sequential updates, modeling temporal evolution through incremental changes to a base model, enabling efficient streaming via difference-based compression. 
% \textbf{OTNeRF}~\cite{OT-NeRF} enforces temporal consistency by modeling scene dynamics as low-frequency variations in pixel distributions, using sliced-Wasserstein distance as a regularizer. 
% \textbf{STGC-NeRF}~\cite{STGC-NeRF} incorporates spatiotemporal geometric constraints for LiDAR-based reconstruction, leveraging scene flow to establish cross-frame correspondences and improve robustness under sparse observations.

% \textbf{STGC-NeRF}~\cite{STGC-NeRF} addresses the inherent challenges of LiDAR-based dynamic reconstruction by incorporating spatial-temporal geometric consistency constraints, leveraging pre-trained scene flow networks to establish pointwise correspondences between frames and mitigate the sparsity and low-frequency limitations of LiDAR data. Similarly, the optimal transport-based approach in \textbf{OTNeRF}~\cite{OT-NeRF} introduces a theoretically grounded regularizer that treats scene dynamics as low-frequency changes in pixel intensity distributions, constraining temporal evolution through sliced-Wasserstein distance minimization to provide a lightweight, architecture-agnostic solution that avoids expensive geometric priors or external preprocessing models. 

% \subsection{\ziren{xxxx}}

% \subsection{\ziren{xxxx}}

%% file: sec/4_methodology_gs.tex
% \section{Dynamic Scene Reconstruction with GS}
%HERE
\section{DYNAMIC 3D GAUSSIAN SPLATTING}

\label{sec:3DGS}

\begin{figure*}[t]
\centering
\includegraphics[width=\textwidth]{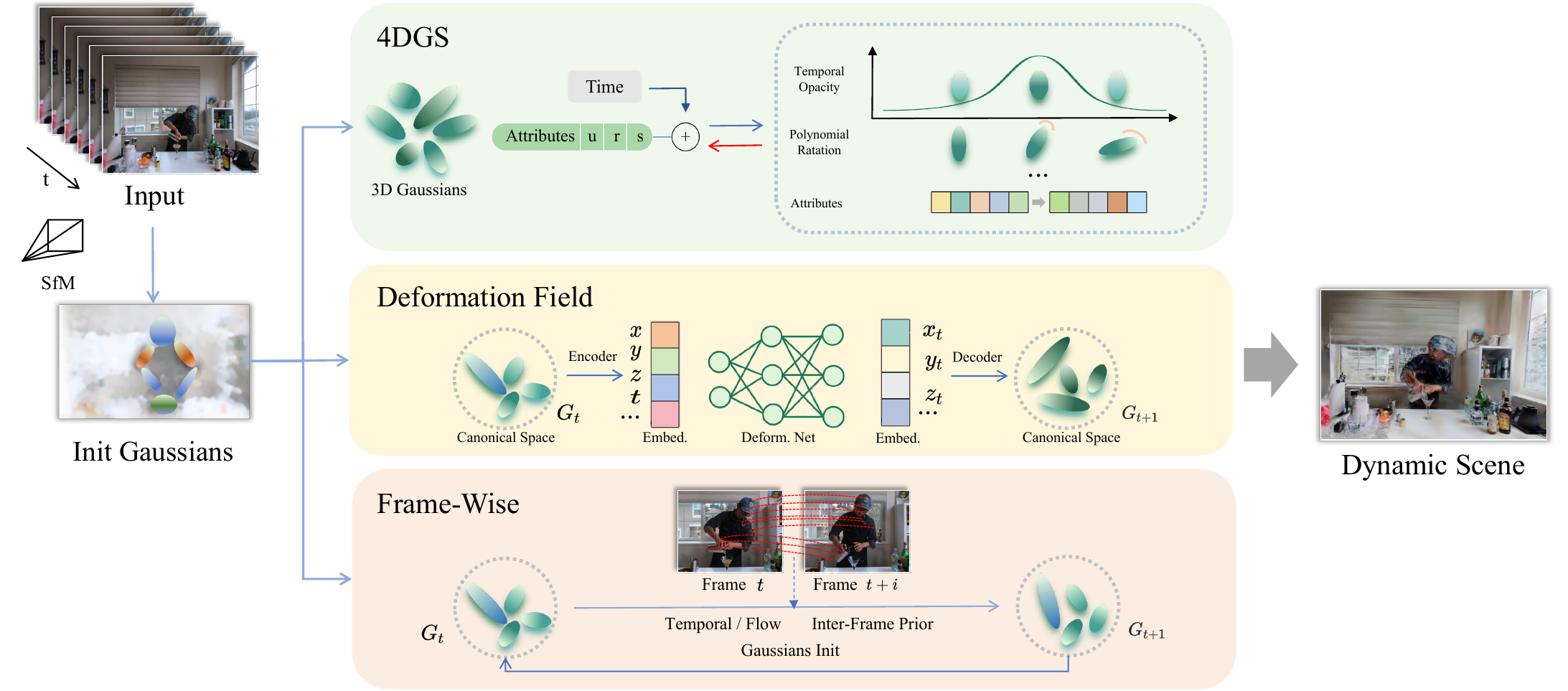}
\vspace{-10pt}
\caption{General pipeline of \textbf{3DGS-style} 4D scene reconstruction methods. The pipeline presents the representative 4D strategies in explicit 4D primitive-based, deformation-field-based, and frame-wise-training frameworks.}
\label{gs_pipeline}
    \vspace{-10pt}
\end{figure*}

% Generally, 4D reconstruction methods based on 3DGS differ fundamentally in how they model temporal dynamics. As illustrated in Fig. \ref{gs_pipeline}, these approaches range from treating time as an explicit dimension to learning implicit deformation fields or performing per-frame optimization. Each strategy presents unique advantages for different scenarios, which we analyze in detail below.

The core challenge in extending 3DGS to the temporal domain lies in parameterizing dynamic scene evolution~\cite{li2026near}. Existing 4D reconstruction frameworks differ in how they model time, ranging from explicit geometric embedding to implicit motion modeling. As shown in Fig.~\ref{gs_pipeline} and Table~\ref{tab:overviewtable}, these approaches can be grouped into three paradigms: (i) \textbf{Explicit 4D Primitive-Based Methods}, which treat time as a geometric dimension; (ii) \textbf{Deformation Field-Based Methods}, which separate geometry and motion via learned transformations; and (iii) \textbf{Frame-wise Training Methods}, which optimize discrete states for temporal consistency. These paradigms involve trade-offs in flexibility, efficiency, and temporal coherence.

\subsection{Fundamentals of 4DGS}

The standard 3DGS framework represents scenes using static primitives, limiting its applicability to stationary environments~\cite{yao2025sd}. Extending to the 4D spatio-temporal domain requires modeling the evolution of scene attributes—position, rotation, and appearance—over time. Existing methods differ in how this temporal evolution is formulated.

\noindent \textbf{Explicit 4D Primitive-Based Methods} treat time as an intrinsic dimension and represent scenes using 4D Gaussian primitives. Each primitive is parameterized by a mean $\boldsymbol{\mu} \in \mathbb{R}^4$ and covariance $\boldsymbol{\Sigma} \in \mathbb{R}^{4 \times 4}$, decomposed into rotation and scaling:
\begin{equation}
    \boldsymbol{\Sigma} = \mathbf{R} \mathbf{S} \mathbf{S}^\top \mathbf{R}^\top, \quad \mathbf{S} = \text{diag}(s_x, s_y, s_z, s_t),
\end{equation}
where the rotation $\mathbf{R}$ couples spatial and temporal dimensions. To render a frame at time $t$, the 4D Gaussian is sliced into a 3D Gaussian via the conditional distribution $p(\mathbf{x}\mid t)$. By partitioning $\boldsymbol{\mu}$ and $\boldsymbol{\Sigma}$ into spatial ($\boldsymbol{\mu}_x, \boldsymbol{\Sigma}_{xx}$), temporal ($\mu_t, \Sigma_{tt}$), and cross terms ($\boldsymbol{\Sigma}_{xt}$), the resulting 3D Gaussian parameters are:
\begin{equation}
    \boldsymbol{\mu}' = \boldsymbol{\mu}_x + \boldsymbol{\Sigma}_{xt}\Sigma_{tt}^{-1}(t - \mu_t), \quad 
    \boldsymbol{\Sigma}' = \boldsymbol{\Sigma}_{xx} - \boldsymbol{\Sigma}_{xt}\Sigma_{tt}^{-1}\boldsymbol{\Sigma}_{xt}^\top.
\end{equation}
This formulation enables integration with standard splatting while maintaining temporal consistency.

\noindent \textbf{Deformation Field-Based Methods} decouple temporal dynamics from scene geometry by maintaining static 3D Gaussians in a canonical space $\mathcal{G}_{\text{can}}$. A learnable deformation network $\mathcal{F}_\theta$ maps canonical coordinates and time to attribute offsets. Given a query point $\mathbf{p}$ and time $t$, the network predicts:
\begin{equation}
    (\Delta \boldsymbol{\mu}, \Delta q, \Delta s) = \mathcal{F}_\theta(\mathbf{p}, t),
\end{equation}
where $\mathbf{p}$ is typically the Gaussian center $\boldsymbol{\mu}$~\cite{Deformable-3D-GS}. The deformed Gaussian set at time $t$ is:
\begin{equation}
    \mathcal{G}(t) = \{ \boldsymbol{\mu} + \Delta \boldsymbol{\mu}, \, q \otimes \Delta q, \, s + \Delta s, \, \alpha, \, c \},
\end{equation}
where $\otimes$ denotes quaternion multiplication. Rendering is performed using standard differentiable splatting.

%HERE
\noindent \textbf{Frame-Wise Training Methods} optimize Gaussian parameters independently at each timestamp, often guided by priors such as rigidity or optical flow. The state of Gaussian $i$ at time $t$ is updated from the previous frame:
\begin{equation}
    \boldsymbol{\mu}_{i,t} = \mathbf{R}_{i,t} \boldsymbol{\mu}_{i,t-1} + \mathbf{T}_{i,t}, \quad 
    q_{i,t} = q(\mathbf{R}_{i,t}) \otimes q_{i,t-1},
\end{equation}
where $\mathbf{R}_{i,t}$ and $\mathbf{T}_{i,t}$ denote local rotation and translation. To handle topology changes, the Gaussian set is dynamically updated:
\begin{equation}
    \mathcal{G}_t = \hat{\mathcal{G}}_t \cup \mathcal{G}_{\text{new}},
\end{equation}
where $\hat{\mathcal{G}}_t$ denotes propagated Gaussians and $\mathcal{G}_{\text{new}}$ newly initialized ones.

\subsection{Explicit 4D Primitive-Based Methods}
These methods extend 3D Gaussian Splatting by incorporating time directly into Gaussian primitives. Combined with temporal slicing and tile-based rasterization, they achieve high reconstruction quality and temporal coherence.

\noindent \textbf{4DGS}~\cite{yang2023gs4d} generalizes 3DGS to 4D by treating time as an additional dimension. It uses 4D scaling and dual quaternions for rotation, and renders by slicing 4D Gaussians into conditional 3D Gaussians. 

Subsequent works refine spatial--temporal disentanglement. \textbf{4D-RotorGS}~\cite{4DRotorGS} adopts rotor-based rotation with entropy regularization, while \textbf{SpaceTimeGS}~\cite{li2024spacetime} models topology changes using temporal basis functions and compact feature decoding.

To better capture motion, \textbf{CD-3DGS}~\cite{CD-3DGS} and \textbf{FreeTimeGS}~\cite{wang2025freetimegs} parameterize motion using Fourier or linear functions, with additional supervision such as optical flow. \textbf{SplatFlow}~\cite{sun2025splatflow} further incorporates learned motion flow fields. 
In order to improve robustness,  \textbf{DriveDreamer4D}~\cite{zhao2025drivedreamer4d} and \textbf{DeSiRe-GS}~\cite{desire-gs} introduce external supervision, including synthetic data and motion-aware decomposition, to handle occlusions and sparse observations.

Explicit 4D primitive-based methods directly embed time into Gaussian parameterization, achieving strong temporal consistency through conditional slicing of 4D Gaussians into 3D counterparts~\cite{yang2023gs4d}. Combined with tile-based rasterization, this formulation enables efficient real-time rendering. Extensions such as temporal basis functions~\cite{li2024spacetime} and rotor-based rotation with entropy regularization~\cite{4DRotorGS} further improve spatio-temporal disentanglement and topology modeling. However, the higher per-primitive parameter dimensionality introduces substantial memory overhead, as each Gaussian requires a 4D mean and a $4\times4$ covariance matrix. These methods are particularly suitable for indoor dynamic scenes with dense multi-view capture and real-time rendering requirements.

\begin{table*}[!t]
\caption{\textbf{Overview of 3DGS-based 4D dynamic scene reconstruction methods.} Methods are categorized into three types. For each method, we summarize its key components and additional priors.}
\label{tab:overviewtable}
\centering
\resizebox{\textwidth}{!}{%
%\rowcolors{2}{gray!15}{white}
\rowcolors{2}{lightcyan}{white}
\begin{tabular}{lcccccccccc}
\hline
\multicolumn{1}{l}{Method} &
  \multicolumn{1}{c}{Venue} &
  \multicolumn{1}{c}{Input} &
  \multicolumn{1}{c}{Scenario} &
  \multicolumn{1}{c}{Target Domain} &
  4D-style &
  \multicolumn{1}{c}{Text} &
  \multicolumn{1}{c}{Flow} &
  Normal &
  Segment. &
  Extra Prior \\ \hline
% \multicolumn{10}{c}{3DGS-Style}                                                                                                                           \\ \hline
4DGS~\cite{yang2023gs4d}           & ICLR2024           & RGB   & Indoor        & Multi.-Centric  & Explicit 4D Primitive  &   &   &   &   &                              \\
CD-3DGS~\cite{CD-3DGS}          & ECCV2024           & RGB   & Indoor        & Entity-Centric  & Explicit 4D Primitive  &   & \checkmark &   &   & RAFT                         \\
SpacetimeGS~\cite{li2024spacetime}      & CVPR2024           & RGB   & Indoor        & Entity-Centric  & Explicit 4D Primitive  &   &   &   &   &                              \\
4DRotorGS~\cite{4DRotorGS}        & ACM SIG. 2024  & RGB   & Indoor        & Entity-Centric  & Explicit 4D Primitive  &   & \checkmark &   &   &                              \\
FreeTimeGS~\cite{wang2025freetimegs}       & CVPR2025           & RGB   & Indoor        & Multi.-Centric  & Explicit 4D Primitive  &   &   &   &   & ROMA                         \\
DeSiRe-GS~\cite{desire-gs}        & CVPR2025           & RGBD & Auto. Driving & Scene-Centric   & Explicit 4D Primitive  &   &   & \checkmark &   &                              \\
SplatFlow~\cite{sun2025splatflow}        & CVPR2025           & RGBD & Auto. Driving & Scene-Centric   & Explicit 4D Primitive  &   & \checkmark  &   & \checkmark  & RAFT                              \\
DriveDreamer4D~\cite{zhao2025drivedreamer4d}   & CVPR2025           & RGBD & Auto. Driving & Scene-Centric   & Explicit 4D Primitive  &   &   &   &   &                              \\
ST-4DGS~\cite{ST-4DGS}          & ACM SIG. 2024  & RGB   & In. \& Wild   & Entity-Centric  & Deformation Fields  &   & \checkmark  &   &   & RAFT                             \\
SaRO-GS~\cite{saro-gs}          & ACM MM2024         & RGB   & Indoor        & Entity-Centric  & Deformation Fields  &   &   &   &   &                              \\
GaussianFlow~\cite{Gaussian-flow}     & CVPR2024           & RGB   & Indoor        & Entity-Centric  & Deformation Fields  &   & \checkmark &   &   & Videoflow                    \\
HUGS~\cite{zhou2024hugs}             & CVPR2024           & RGB   & Auto. Driving & Scene-Centric   & Deformation Fields  &   & \checkmark &   &   & Unimatch                     \\
4D-GS~\cite{wu20244d}            & CVPR2024           & RGB   & Indoor        & Entity-Centric  & Deformation Fields  &   &   &   &   &                              \\
% AYG~\cite{AYG}              & CVPR2024           & indoor    & Deformation Fields  & \checkmark &   &   &   &   & Stable Diffusion\&DM MVDream \\
DeformGS~\cite{DeformGS}         & CVPR2024           & RGB   & Indoor        & Entity-Centric  & Deformation Fields  &   &   &   & \checkmark  &                              \\
SC-GS~\cite{sc-gs}            & CVPR2024           & RGB   & In. \& Wild   & Entity-Centric  & Deformation Fields  &   &   &   &   &                              \\
Deformable-3DGS~\cite{Deformable-3D-GS}  & CVPR2024           & RGB   & Indoor        & Entity-Centric  & Deformation Fields  &   &   &   &   &                              \\
GaGS~\cite{GaGS}             & CVPR2024           & RGB   & In. \& Wild   & Entity-Centric  & Deformation Fields  &   &   &   &   &                              \\
DynMF~\cite{DynMF}            & ECCV2024           & RGB   & Indoor        & Entity-Centric  & Deformation Fields  &   &   &   &   &                              \\
ED-3DGS~\cite{E-D3DGS}          & ECCV2024           & RGB   & In. \& Wild   & Multi.-Centric  & Deformation Fields  &   &   &   &   &                              \\
SwinGS~\cite{SwinGS}           & ECCV2024           & RGB   & Indoor        & Entity-Centric  & Deformation Fields  &   & \checkmark &   &   & RAFT                         \\
GSPrediction~\cite{gaussianprediction}     & ACM SIG.2024       & RGB   & Indoor        & Entity-Centric  & Deformation Fields  &   &   &   &   &                              \\
AmbientGaussian~\cite{AmbientGaussian}  & ACM SIG. 2024      & RGB   & In-the-Wild   & Scene-Centric   & Deformation Fields  &   &   &   &   &                              \\
Marbles~\cite{Marbles}          & SIG-ASIA 2024 & RGB   & Indoor        & Entity-Centric  & Deformation Fields  &   &   &   & \checkmark & Trackanything                \\
% UA-4DGS~\cite{UA-4DGS}          & NIPS2024           & indoor    & Deformation Fields  & \checkmark  &   &   &   & \checkmark & BLIP \& Stable Diffusion     \\
Grid4D~\cite{xu2024grid4d}          & NIPS2024           & RGB   & Indoor        & Entity-Centric  & Deformation Fields  &   &   &   & \checkmark &                              \\
Vidu4D~\cite{wang2024vidu4d}           & NIPS2024           & RGB   & Indoor        & Entity-Centric  & Deformation Fields  & \checkmark &   &   &   &                              \\
4D-GS Wild~\cite{kim20244d}      & NIPS2024           & RGB   & In. \& Wild   & Multi.-Centric  & Deformation Fields  & \checkmark & \checkmark &   &   & RAFT \& BLIP \& Stable Diffusion     \\
HiCoM~\cite{gao2024hicom}            & NIPS2024           & RGB   & Indoor        & Entity-Centric  & Deformation Fields  &   &   &   &   &                              \\
MotionGS~\cite{zhu2024motiongs}         & NIPS2024           & RGB   & Indoor        & Multi.-Centric  & Deformation Fields  &   & \checkmark &   &   & Gaussianflow                 \\
DN-4DGS~\cite{dn-4dgs}          & NIPS2024           & RGB   & In. \& Wild   & Entity-Centric  & Deformation Fields  &   &   &   &   &                              \\
SP-GS~\cite{sp-gs}            & ICML2024           & RGB   & Indoor        & Entity-Centric  & Deformation Fields  &   &   &   &   & SuperPoint                             \\
Gflow~\cite{wang2025gflow}            & AAAI2025           & RGB   & In-the-Wild   & Scene-Centric   & Deformation Fields  &   & \checkmark &   &   & DUSt3R \& UniMatch           \\
EfficientGS~\cite{kong2025efficient}      & AAAI2025           & RGB   & Indoor        & Multi.-Centric  & Deformation Fields  &   & \checkmark &   &   & COLMAP                       \\
% 4D LangSpat~\cite{4d-langsplat}      & CVPR2025           & indoor    & Deformation Fields  & \checkmark &   &   & \checkmark &   & SAM \& CLIP                  \\
Instant GS~\cite{instant-GS}       & CVPR2025           & RGB   & Indoor        & Entity-Centric  & Deformation Fields  &   & \checkmark &   &   & GM-Flow                      \\
BARD-GS~\cite{bard-gs}          & CVPR2025           & RGB   & Indoor        & Entity-Centric  & Deformation Fields  &   &   &   &   &                              \\
MoDec-GS~\cite{MoDec-GS}         & CVPR2025           & RGB   & In. \& Wild   & Multi.-Centric  & Deformation Fields  &   &   &   &   &                              \\
SplineGS~\cite{park2025splinegs}         & CVPR2025           & RGB   & In-the-Wild   & Scene-Centric   & Deformation Fields  &   &   &   &   &                              \\
FreeGave~\cite{li2025freegave}         & CVPR2025           & RGB   & Indoor        & Entity-Centric  & Deformation Fields  &   &   &   & \checkmark & SAM                          \\
GIFStream~\cite{li2025gifstream}        & CVPR2025           & RGB   & Indoor        & Entity-Centric  & Deformation Fields  &   &   &   &   &                              \\
Instruct-4DGS~\cite{instruct-4dgs}    & CVPR2025           & RGB   & Indoor        & Entity-Centric  & Deformation Fields  &  \checkmark &   &   &   &  InstructPix2Pix                            \\
MoSca~\cite{lei2025mosca}            & CVPR2025           & RGB   & In-the-Wild   & Entity-Centric   & Deformation Fields  &   & \checkmark &   &   & RAFT                         \\
TaylorGaussian~\cite{hu2025learnable} & CVPR2025           & RGB   & Indoor        & Entity-Centric  & Deformation Fields  &   &   &   &   &                              \\
SpectroMotion~\cite{spectromotion}    & CVPR2025           & RGB   & Indoor        & Entity-Centric  & Deformation Fields  &   &   & \checkmark &   &                              \\
MoDGS~\cite{qingming2025modgs}            & ICLR2025           & RGBD  & Indoor        & Entity-Centric  & Deformation Fields  &   & \checkmark &   & \checkmark  & RAFT \& SAM2                 \\
% Shape of Motion~\cite{shape_of_motion}  & ICCV2025           & outdoor   & Deformation Fields  &   &   &   &   &   &                              \\
% MonoFusion~\cite{wang2025monofusion}       & ICCV2025           & indoor    & Deformation Fields  &   &   &   &   &   &                              \\
ADC-GS~\cite{ADC-GS}           & IJCAI2025          & RGB   & In. \& Wild   & Entity-Centric  & Deformation Fields  &   &   &   &   &                              \\
Omnire~\cite{chen2024omnire}           & ICLR2025           & RGBD & Auto. Driving & Scene-Centric   & Deformation Fields  &   &   &   &   &                              \\
% \textcolor{yellow}{Dynamic 3D-GS} ~\cite{luiten2024dynamic}   & 3DV2024    & indoor  & Frame-wise training  &   & \checkmark  &   &   &   &                           \\
D-3DG~\cite{luiten2024dynamic}            & 3DV2024            & RGB   & Indoor        & Entity-Centric  & Frame-wise training &   &   &   &   &                              \\
3DGStream~\cite{3dgstream}        & CVPR2024           & RGB   & Indoor        & Entity-Centric  & Frame-wise training &   &   &   &   &                              \\
NPGs~\cite{NPGs}             & CVPR2024           & RGB   & Indoor        & Entity-Centric  & Frame-wise training &   &   &   &   &                              \\
StreetGaussian~\cite{StreetGaussian}   & ECCV2024           & RGBD & Auto. Driving & Scene-Centric   & Frame-wise training &   &   &   & \checkmark & Video K-Net                  \\
DrivingGaussian~\cite{zhou2024drivinggaussian}  & CVPR2024           & RGBD & Auto. Driving & Scene-Centric   & Frame-wise training  &   &   &   &   & \\
Casual-FVS~\cite{Casual-FVS}       & ECCV2024           & RGB   & In-the-Wild   & Scene-Centric   & Frame-wise training &   & \checkmark &   & \checkmark  & RAFT \& SAM                  \\
Ex4DGS~\cite{Ex4DGS}           & NIPS2024           & RGB   & Indoor        & Entity-Centric  & Frame-wise training     &   &   &   &   &                       \\
4DGC~\cite{hu20254dgc}             & CVPR2025           & RGB   & Indoor        & Entity-Centric  & Frame-wise training &   &   &   &   &                              \\
4D-Fly~\cite{4d-fly}           & CVPR2025           & RGB   & Indoor        & Entity-Centric  & Frame-wise training &   &   &   &   &                              \\
MAGS~\cite{MAGS}             & TCSVT2025          & RGB   & Indoor        & Entity-Centric  & Frame-wise training &   & \checkmark &   &   & RAFT                         \\

% VoxelSplat~\cite{zhu2025voxelsplat}       & CVPR2025           & outdoor   & Autonomous Driving  &   &   &   & \checkmark &   &                              \\
% SplatAD~\cite{hess2025splatad}          & CVPR2025           & outdoor   & Autonomous Driving  &   &   &   &   &   &                              \\ 
\hline
% EMD~\cite{wei2024emd}              & ICCV2025           & outdoor   & Autonomous Driving  &   &   &   &   &   &                              \\ 
\end{tabular}%
}
\end{table*}
% \vspace{-10pt}

\subsection{Deformation-Field-Based Methods}
Deformation field-based methods model dynamics by applying a learnable deformation field to a canonical set of 3D Gaussians, typically implemented as an MLP. This approach decouples appearance from motion by predicting only the temporal changes in position, rotation, and scale, while assuming other Gaussian attributes remain fixed.

\textbf{Deformable 3D-GS}~\cite{Deformable-3D-GS} introduces deformation fields into 3DGS by mapping Gaussians to a canonical space and using an MLP to predict position, scale, and rotation offsets, with annealing to reduce rendering jitter. Subsequent works improve efficiency and structural consistency through sparse control. \textbf{SC-GS}~\cite{sc-gs} and \textbf{SP-GS}~\cite{sp-gs} drive dense Gaussians using sparse control points and superpoints, respectively, while \textbf{GaussianPrediction}~\cite{gaussianprediction} employs graph convolutional networks on clustered key points for motion prediction. Feature encoding is further optimized by \textbf{4D-GS}~\cite{wu20244d}, which uses K-Planes with voxel encoding, and \textbf{GaGS}~\cite{GaGS}, which combines point-based MLPs with voxel U-Nets for geometry-aware features.

To reduce artifacts from global deformation, several methods separate static and dynamic components. \textbf{GauFRe}~\cite{liang2025gaufre} and \textbf{SWinGS}~\cite{SwinGS} restrict deformation to dynamic regions. \textbf{HUGS}~\cite{zhou2024hugs} models static backgrounds and dynamic objects with separate parameterizations. \textbf{Gflow}~\cite{wang2025gflow} and \textbf{EfficientGS}~\cite{liu2025efficientgs} further incorporate priors such as depth and optical flow to localize deformable regions.

Hierarchical methods address multi-scale dynamics.
\textbf{Grid4D}~\cite{xu2024grid4d} decomposes spatiotemporal encoding using hash grids and attention. \textbf{MoDec-GS}~\cite{MoDec-GS} and \textbf{Hicom}~\cite{gao2024hicom} adopt global-to-local cascades, while \textbf{ADC-GS}~\cite{ADC-GS} uses anchor-driven deformation to combine transformations with local refinement.

In addition, external priors and structured models are introduced to regularize deformation.
\textbf{MotionGS}~\cite{zhu2024motiongs} and \textbf{MoDGS}~\cite{qingming2025modgs} use optical flow decomposition. \textbf{BARD-GS}~\cite{bard-gs} and \textbf{4D-GS Wild}~\cite{kim20244d} address challenging scenarios with pose interpolation and diffusion-based regularization. \textbf{TaylorGaussian}~\cite{hu2025learnable}, \textbf{Gaussian-Flow}~\cite{Gaussian-flow}, and \textbf{SplineGS}~\cite{park2025splinegs} model motion using analytic representations, while \textbf{FreeGave}~\cite{li2025freegave} enforces physical constraints.

Specialized designs target specific scenarios. \textbf{SpectroMotion}~\cite{spectromotion} handles specular materials, while \textbf{GIFStream}~\cite{li2025gifstream} and \textbf{MoSca}~\cite{lei2025mosca} enable efficient streaming. \textbf{Marbles}~\cite{Marbles} and \textbf{MonoFusion}~\cite{wang2025monofusion} address monocular settings with simplified primitives and motion models. \textbf{OmniRe}~\cite{chen2024omnire} adopts neural scene graphs to separate static and dynamic components under a shared deformation field.

Deformation-field-based methods decouple temporal dynamics from canonical geometry by predicting per-Gaussian attribute offsets through a learnable deformation network~\cite{Deformable-3D-GS}, offering improved parameter efficiency over explicit 4D primitives. This formulation is particularly effective for modeling non-rigid motion and view-dependent specularities. However, MLP-based deformation prediction may introduce rendering jitter, while the additional network inference can limit rendering speed. As one of the most widely explored paradigms, deformation-field-based methods are well-suited for scenes with non-rigid motion and complex appearance changes, and naturally support downstream tasks such as scene editing~\cite{sc-gs,instruct-4dgs} and neural scene graph decomposition~\cite{chen2024omnire}.

\subsection{Frame-Wise Training Methods}

Frame-wise methods optimize 3D Gaussians independently at each timestamp via per-frame reconstruction, optionally incorporating inter-frame constraints for temporal consistency. These approaches are simple and flexible but often incur high storage cost and limited long-term coherence.

\noindent \textbf{D-3DG}~\cite{luiten2024dynamic} models Gaussian centers and orientations as time-varying, while keeping color, opacity, and scale fixed. Motion is guided by rigidity and rotation priors. However, independent optimization can lead to weak temporal consistency and high storage overhead.

To reduce redundancy in complex scenes, many methods adopt dynamic--static decomposition. \textbf{Street Gaussians}~\cite{StreetGaussian} and \textbf{DrivingGaussian}~\cite{zhou2024drivinggaussian} use graph-based representations to separate static backgrounds and dynamic objects~\cite{chen2025gs}. \textbf{Casual-FVS}~\cite{Casual-FVS} decomposes scenes into static planes and dynamic points with flow-based blending, while \textbf{LiveSplats}~\cite{huang2025echoes} employs hierarchical optimization for real-time processing.

For online scenarios with topology changes, \textbf{3DGStream}~\cite{3dgstream} introduces a Neural Transformation Cache to transform existing Gaussians and incrementally add new ones. \textbf{4D-Fly}~\cite{4d-fly} propagates Gaussians across frames using anchor-based updates, expanding the representation only when needed.

In order to reduce storage, interpolation-based methods use keyframes. \textbf{Ex4DGS}~\cite{Ex4DGS} and \textbf{4DGC}~\cite{hu20254dgc} reconstruct intermediate frames via interpolation or motion prediction. \textbf{Ex4DGS} applies CHip and Slerp for trajectory smoothing, while \textbf{4DGC} uses multi-resolution motion grids for efficient transformation estimation.
On the other hand, several methods incorporate strong supervision to enforce temporal consistency. 
\textbf{GaussianFlow}~\cite{gao2024gaussianflow} enforces consistency between 3D motion and 2D observations via optical flow. \textbf{MAGS}~\cite{MAGS} further improve supervision using dense correspondences and uncertainty-aware flow modeling.

Frame-wise methods provide architectural simplicity and flexibility by optimizing 3D Gaussians independently at each timestamp, with dynamic set updates ($\mathcal{G}_t = \hat{\mathcal{G}}*t \cup \mathcal{G}*{\text{new}}$) that naturally handle topology changes~\cite{3dgstream,4d-fly}. This design is particularly suitable for online and streaming scenarios. However, the independent per-frame parameterization results in storage costs that grow linearly with sequence length, while long-range temporal coherence remains limited without explicit inter-frame coupling. Keyframe interpolation~\cite{Ex4DGS,hu20254dgc} and flow-supervised consistency losses~\cite{MAGS} partially alleviate these issues, but introduce approximation errors or reliance on external priors. Consequently, this paradigm is best suited for short sequences, casual monocular videos~\cite{Casual-FVS}, and streaming reconstruction tasks where real-time adaptability is prioritized over long-term temporal consistency~\cite{shao2025eventvad,shao2024accidentblip}.

%% file: sec/5_dataset_and_benchmark.tex
\section{Performance Evaluation}
In this section, we summarize representative datasets for dynamic scene synthesis, categorizing them according to key properties and research objectives (Table~\ref{tab:dynamic_scene_datasets}). In addition, we explore novel view synthesis  and geometric reconstruction in representative benchmarks, highlighting the best results as \colorbox[HTML]{BDE6CD}{\textbf{first}}, \colorbox[HTML]{E2EEBC}{second}, and \colorbox[HTML]{FFF3BB}{third}. We organize quantitative data from papers with a common evaluation protocol and cross-verified results. Since some works do not release codes or specific configurations, our priority is to include papers with consistent benchmarks, ensuring a reliable basis for verifiable comparison with a shared evaluation framework across multiple sources.
% \yan{We summarize representative datasets for dynamic scene synthesis. Rather than organizing them by environment type, we categorize them according to key benchmark properties and research objectives, including sensor setup, temporal scale, and scene complexity. Table~\ref{tab:dynamic_scene_datasets} provides an overview of these attributes.}

\begin{table*}[htbp]
    \centering
    \caption{\textbf{Taxonomy of dynamic scene datasets based on benchmark properties.} Datasets are grouped by their primary research focus and capture characteristics.}
    \label{tab:dynamic_scene_datasets}
    \resizebox{\textwidth}{!}{
    \rowcolors{2}{lightcyan}{white}
    \begin{tabular}{lcccccc}
    \hline
    \textbf{Dataset} & \textbf{Scene Type} & \textbf{Sensor Setup} & \textbf{Resolution} & \textbf{Frame Rate} & \textbf{Scene/Seq} & \textbf{Temporal Scale} \\
    \hline
    \rowcolor{gray!20} \multicolumn{7}{l}{\textit{Synthetic Datasets (Ground Truth Geometry/Motion)}} \\
    D-NeRF~\cite{li2021dnerf} & Indoor & 1 Cam & 800$\times$800 & -- & 8 & 50--200 frames \\
    ParticleNeRF~\cite{abou2024particlenerf} & Indoor & 40 Cams & -- & -- & 6 & -- \\
    SS3DM~\cite{hu2024ss3dm} & Autonomous Driving & 6 Cams + 5 LiDAR & -- & 10 FPS & 28 & 13K frames \\
    
    \hline
    \rowcolor{gray!20} \multicolumn{7}{l}{\textit{Real-world: Monocular \& Sparse View}} \\
    DAVIS~\cite{pont20172017} & Outdoor & 1 Cam & -- & -- & 150 & 10k frames total \\
    HyperNeRF~\cite{park2021hypernerf} & Indoor/Outdoor & 1--2 Cams & 540$\times$960 & 15 FPS & 17 & 8--15s/seq \\
    DyCheck~\cite{gao2022monocular} & Indoor & 1 iPhone + 7 Static & -- & -- & 14 & 200--500 frames \\
    Stereo4D~\cite{jin2024stereo4d} & Indoor/Outdoor & 2 Cams & Diverse & -- & 200K clips & -- \\
    NeRF-DS~\cite{NeRF-DS} & Outdoor & 2 Cams & -- & -- & 8 & -- \\
    
    \hline
    \rowcolor{gray!20} \multicolumn{7}{l}{\textit{Real-world: Dense Multi-view \& Human-Centric}} \\
    Panoptic Studio~\cite{joo2015panoptic} &  Indoor & 480 Cams & -- & -- & 5 & -- \\
    ENeRF-Outdoor~\cite{lin2022efficient} & Outdoor & 18 Cams & -- & -- & 3 & 1200 frames \\
    Neu3DV~\cite{li2022neural} & Indoor & 18--21 Cams & 2704$\times$2028 & 30 FPS & 6 & 10s/seq \\
    Technicolor~\cite{Technicolor} & Indoor (RGB-only) & 16 Cams & 2048$\times$1088 & 25 FPS & 12 & -- \\
    NVIDIA Dynamic~\cite{yoon2020novel} & Outdoor & 12 Cams & -- & -- & 7 & 90--200 frames \\
    Meeting Room~\cite{meetingroomdataset} & Indoor & 13 Cams & 1280$\times$720 & 30 FPS & 3 & 300 frames \\
    Google Immersive~\cite{broxton2020immersive} & Indoor/Outdoor & $\leq$46 Cams & -- & -- & 15 & -- \\
    
    \hline
    \rowcolor{gray!20} \multicolumn{7}{l}{\textit{Real-world: Long Horizon \& Multimodal}} \\
    Waymo Open~\cite{sun2020scalability} & Autonomous Driving & 5 Cams + 5 LiDAR & 1920$\times$1280 & 10 FPS & 1150 & $\sim$12M frames \\
    nuScenes~\cite{caesar2020nuscenes} & Autonomous Driving & 6 Cams + LiDAR & 1600$\times$900 & 12 FPS & 1000 & $\sim$5.5h \\
    Argoverse 2~\cite{Argoverse2} & Autonomous Driving & 7 Cams + LiDAR & 1920$\times$1200 & 30 FPS & 1000 & $\sim$1000h \\
    PandaSet~\cite{xiao2021pandaset} & Autonomous Driving & 6 Cams + LiDAR & 1920$\times$1080 & 20 FPS & 103 & $\sim$1h \\
    OmniHD-Scenes~\cite{zheng2024omnihd} & Autonomous Driving & 6C + LiDAR + 6R & Diverse & 15 FPS & 1501 & $\sim$30s/seq \\
    KITTI~\cite{geiger2013vision} & Autonomous Driving & 2 Stereo + LiDAR & 1242$\times$375 & 10 FPS & 22 & $\sim$6h \\
    WayveScenes101~\cite{zurn2024wayvescenes101} & Autonomous Driving (RGB-only) & 5 Cams & -- & 10 FPS & 101 & 20s/seq \\
    \hline
    \end{tabular}
    }
\end{table*}

\subsection{Benchmark Datasets}

\subsubsection{Synthetic vs. Real-world Data}

\noindent \textbf{Synthetic:} \textbf{D-NeRF}~\cite{li2021dnerf} and \textbf{ParticleNeRF}~\cite{abou2024particlenerf} provide clean articulated and deformable scenes, making them suitable for evaluating deformation modeling and motion tracking. In autonomous driving, \textbf{SS3DM}~\cite{hu2024ss3dm} offers synchronized RGB, LiDAR, and semantic annotations, enabling controlled evaluation of multimodal fusion methods. However, limited domain diversity and simplified rendering reduce their ability to assess real-world robustness.

\noindent \textbf{Real-world:} \textbf{HyperNeRF}~\cite{park2021hypernerf} and \textbf{Technicolor}~\cite{Technicolor} introduce complex lighting, calibration errors, and dynamic backgrounds. These datasets are more suitable for evaluating generalization and robustness, but often lack accurate ground truth, making quantitative evaluation more challenging.

\subsubsection{Monocular vs. Multi-view Capture}

\noindent \textbf{Monocular:} \textbf{D-NeRF}~\cite{li2021dnerf}, \textbf{HyperNeRF}~\cite{park2021hypernerf}, and \textbf{DAVIS}~\cite{pont20172017} contain sequences captured from a single moving camera. These benchmarks are well-suited for evaluating methods that rely on strong priors, such as generative models or deformation-aware representations. However, monocular setups often suffer from scale ambiguity and limited spatial coverage.

\noindent \textbf{Multi-view:} \textbf{Panoptic Studio}~\cite{joo2015panoptic} and \textbf{Google Immersive}~\cite{broxton2020immersive} provide dense spatial coverage, making them suitable for evaluating reconstruction fidelity and view consistency. Intermediate-scale datasets such as \textbf{NeRF-DS}~\cite{NeRF-DS} and \textbf{KITTI}~\cite{geiger2013vision} instead offer sparse multi-view setups that better reflect real-world constraints and are useful for evaluating view-sparse reconstruction methods.

\subsubsection{Short-term vs. Long-term Horizons}

\noindent \textbf{Short Horizon:} \textbf{Neu3DV}~\cite{li2022neural} and \textbf{Meeting Room}~\cite{meetingroomdataset} typically contain short clips with localized motions. These datasets are well-suited for evaluating deformation modeling and short-term motion consistency, but may not capture long-range dynamics.

\noindent \textbf{Long Horizon:} \textbf{Argoverse 2}~\cite{Argoverse2} and \textbf{nuScenes}~\cite{caesar2020nuscenes} provide large-scale temporal data across diverse driving environments. These benchmarks are useful for evaluating temporal consistency and long-term prediction, although sparse viewpoints and noisy annotations introduce additional challenges.

\subsubsection{Human-Centric vs. General Dynamic Scenes}

\noindent \textbf{Human-Centric:} \textbf{Panoptic Studio}~\cite{joo2015panoptic} and \textbf{ENeRF-Outdoor}~\cite{lin2022efficient} focus on articulated human motion. These datasets are particularly suitable for evaluating deformation modeling, skeletal motion tracking, and fine-grained geometry reconstruction.

\noindent \textbf{General Dynamics:} \textbf{NVIDIA Dynamic Scene}~\cite{yoon2020novel} and \textbf{Stereo4D}~\cite{jin2024stereo4d} include diverse dynamic elements such as animals, fluids, and object interactions. These datasets better reflect real-world complexity, but introduce greater challenges for motion decomposition and scene understanding.

\subsubsection{RGB-only vs. Multimodal Datasets}

\noindent \textbf{RGB-only:} \textbf{Technicolor}~\cite{Technicolor} and \textbf{WayveScenes101}~\cite{zurn2024wayvescenes101} rely solely on visual inputs. These benchmarks are suitable for evaluating appearance modeling, but suffer from depth ambiguity and sensitivity to lighting variations.

\noindent \textbf{Multimodal:} \textbf{Waymo Open Dataset}~\cite{sun2020scalability}, \textbf{PandaSet}~\cite{xiao2021pandaset}, and \textbf{OmniHD-Scenes}~\cite{zheng2024omnihd} integrate LiDAR, radar, and IMU signals, making them suitable for evaluating geometric accuracy and robust reconstruction under challenging conditions. \textbf{DyCheck}~\cite{gao2022monocular} further incorporates smartphone LiDAR, enabling evaluation in lightweight capture settings.

\begin{figure*}[tp]
\centering
\includegraphics[width=\textwidth]{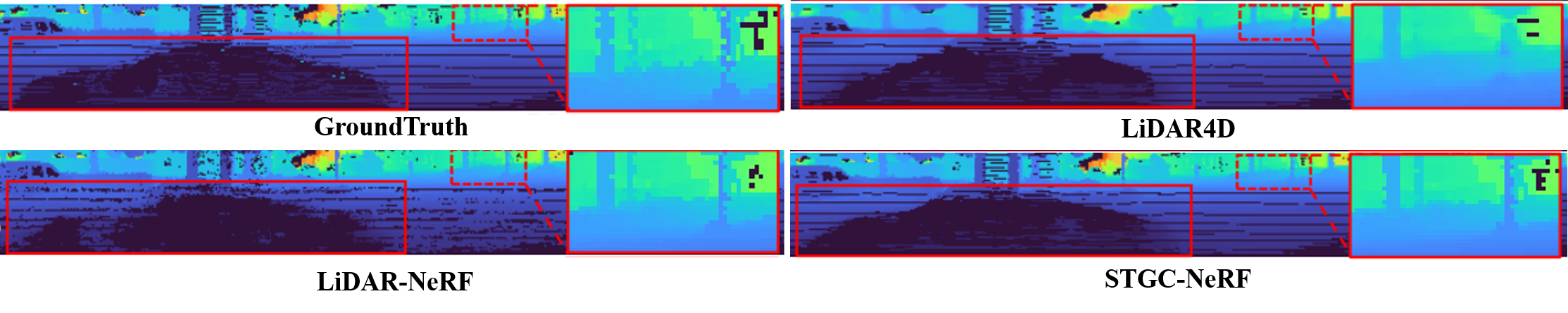}
\vspace{-15pt}
\caption{\textbf{Qualitative reconstruction point map and depth of NeRF-style methods on the NuScenes~\cite{caesar2020nuscenes} dataset.} Image from~\cite{STGC-NeRF}.}
\label{nerf_nuscenes}
\vspace{-10pt}
\end{figure*}

\subsection{Evaluation Metrics}
% Evaluation of 4D reconstruction considers rendering quality, geometric accuracy, and efficiency. We adopt standard metrics for each aspect.

% \begin{itemize}
% \item \textbf{PSNR} measures reconstruction fidelity via pixel-wise error.
% \item \textbf{SSIM} evaluates perceptual similarity in structure and contrast.
% \item \textbf{LPIPS} measures perceptual similarity using deep features.
% \item \textbf{CD} measures geometric similarity between point sets.
% \item \textbf{F-Score} evaluates reconstruction accuracy via precision and recall.
% \item \textbf{RMSE} measures error between predicted and ground-truth geometry.
% \item \textbf{FPS} measures inference speed.
% \end{itemize}

Evaluation of 4D reconstruction involves three primary aspects: rendering quality, geometric accuracy, and computational efficiency. We adopt widely used metrics in the literature and further analyze their strengths and limitations for dynamic 4D reconstruction.

%\begin{itemize}
\subsubsection{View Synthesis Metrics}

\noindent \textbf{PSNR} measures reconstruction fidelity through pixel-wise error. Although widely used for novel view synthesis, it primarily evaluates per-frame appearance quality and does not fully capture temporal consistency in dynamic scenes.

\noindent \textbf{SSIM} evaluates perceptual similarity in terms of structure and contrast. Compared to PSNR, it better reflects structural preservation, but remains a frame-wise metric without explicit temporal modeling.

\noindent \textbf{LPIPS} measures perceptual similarity using deep feature representations. While it correlates better with human perception, it mainly evaluates appearance quality rather than dynamic geometric accuracy.

\subsubsection{Geometric and Spatiotemporal Metrics}

\noindent \textbf{Chamfer Distance (CD)} measures geometric similarity between predicted and ground-truth point sets. It is widely used to evaluate geometric fidelity, but remains sensitive to point density and may not adequately capture topology changes in dynamic scenes.

\noindent \textbf{F-Score} evaluates reconstruction quality through precision and recall under a distance threshold. It provides a more balanced assessment of geometric accuracy, although the results can vary with threshold selection.

\noindent \textbf{RMSE} measures the error between predicted and ground-truth geometry. While it offers a direct measure of geometric accuracy, it is sensitive to outliers and may not fully reflect perceptual quality.

\subsubsection{Efficiency and Practicality}

\noindent \textbf{FPS} measures inference speed and computational efficiency. However, FPS alone does not capture other practical factors such as memory consumption and scalability, which are also critical in 4D reconstruction.

%\end{itemize}

Despite these limitations, these metrics remain widely adopted in existing 4D reconstruction works and provide a common basis for comparison. However, the lack of standardized protocols for evaluating temporal coherence, topology changes, and long-term consistency remains an open challenge in dynamic 4D reconstruction.

\subsection{Novel View Synthesis}

We evaluate rendering fidelity using PSNR, D-SSIM, SSIM, and LPIPS. Benchmarks are conducted on widely used datasets, including Neu3D~\cite{li2022neural}, D-NeRF~\cite{li2021dnerf}, NeRF-DS~\cite{NeRF-DS}, NVIDIA Dynamic Scene~\cite{yoon2020novel}, and Waymo~\cite{sun2020scalability}.

\begin{table}[tp]
\centering
\footnotesize
%\rowcolors{2}{gray!15}{white}
\rowcolors{2}{lightcyan}{white}
\caption{\textbf{Neu3D~\cite{li2022neural} NeRF-style 4D reconstruction results.} PSNR ($\uparrow$), SSIM ($\uparrow$), and LPIPS ($\downarrow$) are used as metrics. }
\label{tab:dynerf}
% \resizebox{\columnwidth}{!}{%
\begin{tabular}{lccc}
\hline
Methods         & PSNR ($\uparrow  $)  & SSIM ($\uparrow  $) & LPIPS ($\downarrow $) \\ \hline
DyNeRF         & 29.6  & \snd0.961 & 0.083 \\
StreamRF       & 28.3     & -     & -      \\
HexPlane       & 29.5  & -     & 0.097  \\
K-Planes       & 31.6  & \fst\textbf{0.964} & -      \\
TIDNeRF        & 29.9      & -     & 0.096  \\
HyperReel      & 31.1   & 0.927 & 0.096  \\
MixVoxels      & 31.7  & \trd0.944 & \trd0.064  \\
MSTH           & \fst\textbf{32.4}  & -    & \fst\textbf{0.056}  \\
NeRFPlayer     & 30.7   & 0.931 & 0.111  \\
Sync-NeRF      & \trd31.9    & 0.916 & 0.146  \\
DecouplingNeRF & 28.6     & 0.917 & 0.123  \\
Ced-NeRF       & 30.6     & 0.919 & -      \\
Gear-NeRF      & 31.8     & 0.936 & \snd0.058  \\
DaReNeRF       & \snd32.3    & -     & 0.084  \\ \hline
\end{tabular}%
% }
\vspace{-10pt}
\end{table}

\begin{table}[tp]
\centering
\footnotesize
%\rowcolors{2}{gray!15}{white}
\rowcolors{2}{lightcyan}{white}
\caption{\textbf{Neu3D~\cite{li2022neural} 3DGS-style 4D reconstruction results.}
PSNR ($\uparrow  $), SSIM ($\uparrow  $), and LPIPS ($\downarrow $) are used as the evaluation metrics.}
\label{tab:hypernerf}
\begin{tabular}{lccc}
\hline
               & PSNR ($\uparrow  $) & SSIM ($\uparrow  $) & LPIPS ($\downarrow $) \\ \hline
% GaussianFlow  & 32.30  & -     & -      & - \\
% 4D-GS         & 31.15  & 0.016 & -      & 0.049 \\
4DGS          & 32.01  & -      & 0.055 \\
4DRotorGS     & 31.62  & 0.940  & 0.140      \\
FreeTimeGS    & \fst\textbf{33.19}  & -      & \fst\textbf{0.036 } \\
SpacetimeGS   & 32.05  & -      & 0.044  \\
CD-3DGS       & 30.46  & \trd0.955 & 0.150  \\%
% Grid4D        & 31.49  & 0.936  & -  \\
SaRO-GS       & 32.15  & -      & 0.044  \\
ST-4DGS       & \trd32.67  & 0.946  & 0.166  \\
% IGS           & 32.14  & 0.011 & -      & 0.039  \\
% 3DGStream     & 27.03  & \textbf{0.895} & -     & -  \\
4DGC          & 31.58  & 0.943  & -  \\
ADC-GS        & 31.67  & \fst\textbf{0.981} & 0.061  \\
GIFStream     & 31.75  & 0.938 & 0.051  \\ %
% Instruct-4DGS & 22.22  & -     & 0.609 & 0.330  \\
TaylorGaussian& \snd33.02  & \snd0.970 & 0.053  \\
GaGS          & 31.31  & 0.950 & 0.140 \\
DynMF         & 31.70  & 0.946 & 0.180  \\
DN-4DGS       & 32.02  & 0.944 & \trd0.043      \\
ED-3DGS       & 31.31  & 0.945 & \snd0.037    \\ 
Ex4DGS        & 32.11  & \snd0.970 & 0.048  \\
MAGS          & 31.30  & 0.943 & 0.053      \\
% HiCoM         & 31.00  & -     & - & -  \\
% SwinGS        & 31.47  & -     & -     & -      \\
% MoDGS         & 22.64  & -     & 0.804 & 0.155    \\ 
\hline
\end{tabular}
\vspace{-10pt}
\end{table}

\noindent \textbf{Neu3D}. Table~\ref{tab:dynerf} reports NeRF-style results under the protocol of~\cite{li2022neural}. Performance steadily improves from early implicit models to recent hybrid approaches. DyNeRF establishes a strong baseline, while MSTH and DaReNeRF further advance the state of the art. A notable trend is the strong performance of 4D feature-volume-based architectures, which consistently achieve leading results across multiple metrics. In addition, specialized designs highlight the benefits of structured priors; for example, the uncertainty-aware modeling in MSTH and the semantic segmentation constraints in Gear-NeRF demonstrate the effectiveness of incorporating probabilistic cues and geometric semantics into dynamic scene optimization.
% with MSTH leading in PSNR and LPIPS due to its multi-scale temporal modeling.

\begin{figure*}[t]
\centering
\includegraphics[width=\textwidth]{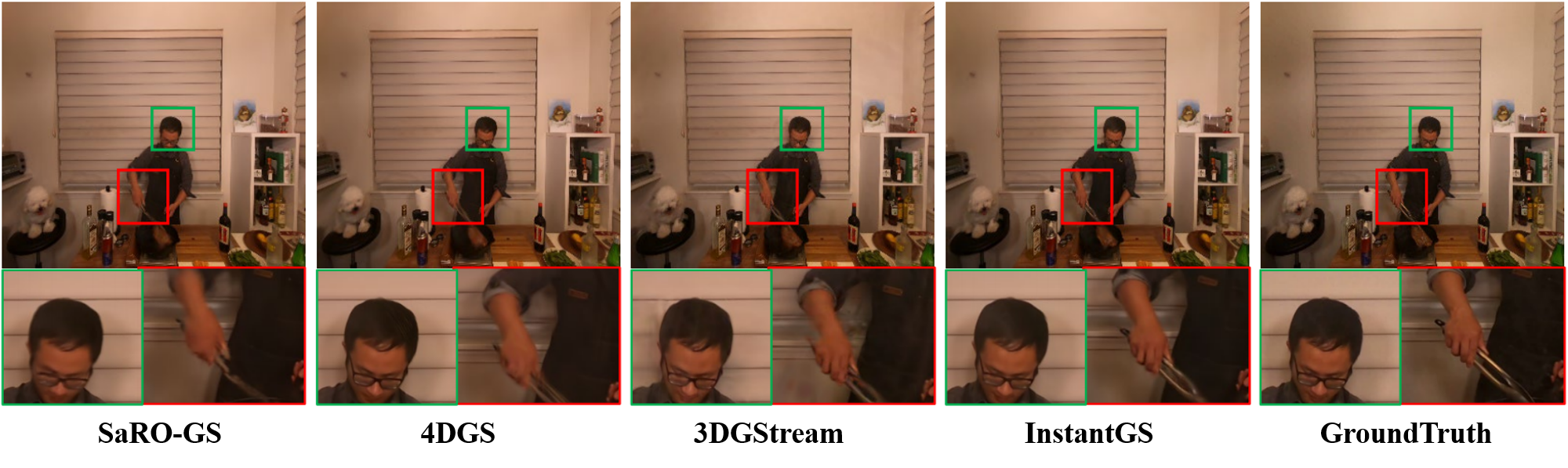}
    \vspace{-10pt}
\caption{\textbf{Qualitative Novel View Synthesis results of 3DGS-style framework on the Neu3D \cite{li2022neural} dataset.} Image sourced from \cite{instant-GS}.}
\label{NVS_results}
    \vspace{-10pt}
\end{figure*}

Table~\ref{tab:hypernerf} presents results for 3DGS-based methods. Recent approaches, such as FreeTimeGS and TaylorGaussian, outperform earlier methods in PSNR while maintaining strong perceptual quality. Deformation-based methods also perform competitively: ADC-GS achieves the best SSIM, and ED-3DGS yields strong LPIPS scores. A key observation from the evaluation is the performance improvement enabled by integrating external priors. In addition, specialized frameworks highlight the effectiveness of structured constraints; for example, the feature matching priors in FreeTimeGS, which achieves high rendering fidelity, and the optical flow constraints used in ST-4DGS and CD-3DGS demonstrate the benefits of external guidance for dynamic scene optimization. These priors effectively regularize Gaussian primitives in highly dynamic regions, reducing floaters and multi-view inconsistencies commonly observed in purely photometric optimization.
Figure~\ref{NVS_results} shows qualitative comparisons on Neu3D.

\begin{table}[tp]
\centering
\footnotesize
%\rowcolors{2}{gray!15}{white}
\rowcolors{2}{lightcyan}{white}
\caption{\textbf{D-NeRF~\cite{li2021dnerf} NeRF-style 4D reconstruction results.} PSNR ($\uparrow$), SSIM ($\uparrow$), and LPIPS ($\downarrow$) are used as metrics.}
\label{tab:dnerf_nerf}
\begin{tabular}{lccc}
\hline
Methods  & PSNR ($\uparrow  $)  & SSIM ($\uparrow  $) & LPIPS ($\downarrow $) \\ \hline
D-NeRF   & 30.50  & 0.95 & 0.070 \\
TiNeuVox & 32.67 & \trd0.97 & 0.041 \\
HexPlane & 31.04 & \trd0.97 & 0.040 \\
K-Planes & 31.61 & \trd0.97 & 0.049 \\
TIDNeRF  & \trd32.73 & \trd0.97 & \trd0.033 \\
Ced-NeRF & \snd34.21 & \fst\textbf{0.99} & 0.037 \\
DaReNeRF & 31.95 & \trd0.97 & \snd0.030 \\
SLS4D    & \fst\textbf{34.84} & \snd0.98 & \fst\textbf{0.025} \\ \hline
\end{tabular}%
\vspace{-10pt}
\end{table}

\noindent \textbf{D-NeRF}. Table~\ref{tab:dnerf_nerf} reports reconstruction quality under the protocol of~\cite{li2021dnerf}, showing that 4D feature-volume-based methods consistently achieve state-of-the-art performance across diverse dynamic benchmarks. While earlier approaches such as TiNeuVox establish strong baselines, Ced-NeRF and SLS4D leverage semi-explicit representations and high-dimensional feature volumes to better disentangle static geometry from temporal dynamics. This trend reflects a broader shift in modeling small-scale dynamic indoor scenes, where grid-based feature structures outperform purely coordinate-based MLPs.
% This design mitigates the spectral bias of MLPs and improves reconstruction of high-frequency dynamics.

% \usepackage{graphicx}
\begin{table}[tp]
\centering
\footnotesize
%\rowcolors{2}{gray!15}{white}
\rowcolors{2}{lightcyan}{white}
\caption{\textbf{NeRF-DS~\cite{NeRF-DS} 3DGS-style 4D reconstruction results.} PSNR ($\uparrow$), SSIM ($\uparrow$), and LPIPS ($\downarrow$) are used as metrics.}
\label{tab:nerfds_gs}
\begin{tabular}{lccc}
\hline
Methods & \multicolumn{1}{l}{PSNR ($\uparrow  $)} & \multicolumn{1}{l}{SSIM ($\uparrow  $)} & \multicolumn{1}{l}{LPIPS ($\downarrow $)} \\ \hline
Deformable-3DGS & 24.10  & 0.85 & \trd0.18  \\
SC-GS           & 24.10  & \snd0.89 & \fst\textbf{0.14}   \\
4D-GS           & 24.18 & \trd0.88 & \fst\textbf{0.14} \\
SP-GS           & 23.33 & 0.84 & 0.21  \\
MotionGS        & \snd24.54 & 0.87 & \snd0.17  \\
DN-4DGS         & \trd24.36 & 0.87 & \snd0.17  \\ 
EfficientGS     & \fst\textbf{24.65} & \fst\textbf{0.90}   & \fst\textbf{0.14}  \\\hline
\end{tabular}%
\vspace{-10pt}
\end{table}

\noindent \textbf{NeRF-DS}. Table~\ref{tab:nerfds_gs} reports rendering performance under the protocol of \cite{NeRF-DS}. EfficientGS achieves the best results, while motion-aware methods such as MotionGS and DN-4DGS perform strongly on dynamic regions. This trend highlights the effectiveness of deformation-field-based representations for explicit 4D modeling. By decoupling temporal motion from canonical geometry, these methods avoid the parameter growth associated with unified 4D primitives. Moreover, this formulation is well-suited for modeling complex non-rigid dynamics and view-dependent specularities, which are particularly challenging in the NeRF-DS dataset.

\begin{figure*}[tp]
\centering
\includegraphics[width=\textwidth]{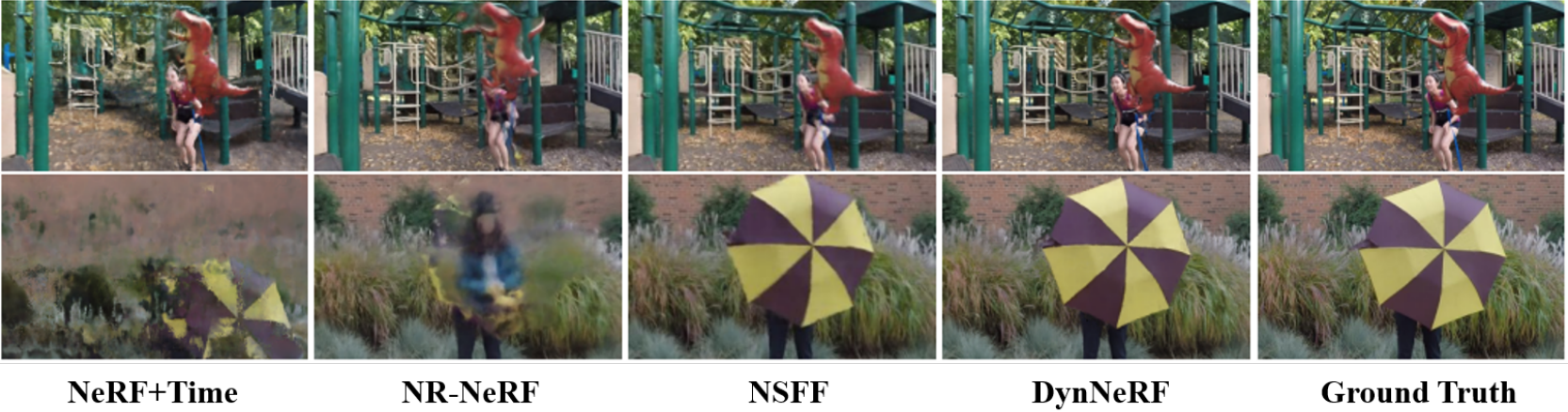}
    \vspace{-10pt}
\caption{\textbf{Qualitative novel view synthesis results of NeRF-style methods on the NVIDIA Dynamic Scene~\cite{yoon2020novel} dataset.} Image from~\cite{dynnerf}.}
\label{nvidiads_vis}
    \vspace{-5pt}
\end{figure*}

\noindent \textbf{NVIDIA Dynamic Scene Dataset}. Qualitative evaluations in unstructured, in-the-wild environments (Fig.~\ref{nvidiads_vis}) highlight the strong performance of frameworks incorporating geometric priors. DynNeRF maintains superior free-view consistency and temporal stability, largely due to its use of multi-view constraints and 3D scene flow for regularization. This demonstrates the effectiveness of geometric priors in dynamic scene reconstruction.
% Figure~\ref{nvidiads_vis} shows qualitative results in unconstrained settings. DynNeRF demonstrates strong free-viewpoint consistency, benefiting from multi-view constraints such as scene flow.

\begin{figure*}[tp]
\centering
\includegraphics[width=\textwidth]{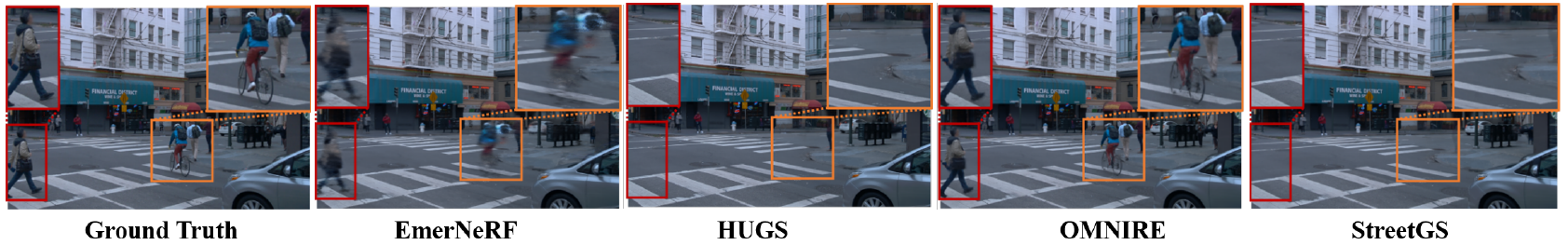}
    \vspace{-10pt}
\caption{\textbf{Qualitative novel view synthesis results of 4D reconstruction methods on the Waymo~\cite{sun2020scalability} dataset.} Image from~\cite{chen2024omnire}.}
\label{waymo_vis}
\vspace{-10pt}
\end{figure*}

\noindent \textbf{Waymo}. As illustrated in Figure~\ref{waymo_vis}, general-purpose 4D reconstruction methods often struggle with distant or fast-moving objects in large-scale driving scenes, leading to ghosting artifacts or geometric collapse. In contrast, EmerNeRF achieves more robust results by integrating 3D scene flow with DINOv2 features, providing a more stable optimization signal and improved robustness to lighting variations. OmniRE further delivers high-fidelity reconstruction by incorporating category-specific semantic priors for human modeling, effectively constraining the solution space to physically plausible structures. These results highlight the importance of semantic guidance and structural priors for 4D reconstruction under limited viewpoint overlap.

\subsection{Geometric Reconstruction}

We evaluate geometric fidelity using surface extraction and point-based metrics, considering both vision-only and vision-LiDAR methods. For image-based methods, we follow the D-NeRF~\cite{li2021dnerf} surface extraction pipeline. For vision--LiDAR methods, we directly compute point cloud error via nearest-neighbor distance. Meshes are extracted from SDF zero-crossings using marching cubes~\cite{marching-cube}. We report Chamfer Distance (CD), RMSE, and F1-score (F1) with a 5\,cm threshold.

\noindent \textbf{NuScenes}. Evaluations on the NuScenes dataset (Table~\ref{tab:nuscenes}) show that frameworks incorporating LiDAR supervision significantly outperform vision-only approaches, highlighting the importance of active depth sensing for resolving scale ambiguities in large-scale environments. STGC-NeRF and LiDAR4D achieve the lowest global geometric error (Fig.~\ref{nerf_nuscenes}) by leveraging spatio-temporal flow and surface normal priors for reconstruction regularization. Furthermore, the 3DGS-based OmniRE yields superior local depth accuracy and robustness to outliers, suggesting that deformation-field-based representations are effective for capturing fine geometric details and preserving structural integrity in dynamic driving scenes.
% Table~\ref{tab:nuscenes} compares implicit (NeRF) and explicit (GS) methods. Methods with LiDAR supervision significantly outperform vision-only approaches. STGC-NeRF achieves the lowest geometric error (Fig.~\ref{nerf_nuscenes}), highlighting the importance of depth sensing. The 3DGS-based OmniRE further improves local depth accuracy and robustness to outliers.

\begin{table}[tp]
\centering
\small
%\rowcolors{2}{gray!15}{white}
\rowcolors{2}{lightcyan}{white}
\caption{\textbf{NuScenes~\cite{caesar2020nuscenes} 3D geometric reconstruction results.} * denotes methods with LiDAR supervision; $\dagger$ uses protocols from~\cite{chen2024omnire}.}
\label{tab:nuscenes}

\begin{tabular}{lccc}
\hline
methods & CD ($\downarrow$) & F-Score ($\uparrow$) & RMSE ($\downarrow$) \\
\hline
\multicolumn{4}{c}{NeRF-style} \\
\hline
D-NeRF          & 0.33 & 0.85 & 7.11 \\
TiNeuVox-B      & 0.39 & 0.86 & 7.21 \\
K-Planes        & \trd0.30 & \trd0.89 & \trd6.80 \\
LiDAR4D*        & \snd0.24 & \snd0.89 & \snd6.78 \\
STGC-NeRF*      & \fst\textbf{0.22} & \fst\textbf{0.91} & \fst\textbf{6.54} \\
\hline
\multicolumn{4}{c}{3DGS-style} \\
\hline
Deformable-3DGS$\dagger$ & \trd0.38 & -    & \trd2.97 \\
StreetGaussian$\dagger$        & \snd0.27 & -    & \snd2.19 \\
OmniRE$\dagger$          & \fst\textbf{0.24} & - & \fst\textbf{1.89} \\
\hline
\end{tabular}
\vspace{-10pt}
\end{table}

\subsection{Model Efficiency}

We evaluate efficiency using GPU memory (peak GB), FPS, and training time on four NVIDIA A100 GPUs. Table~\ref{tab:performance} summarizes results for NeRF-style and 3DGS-style methods on Neu3D.
Among NeRF-style methods, DevRF and StreamRF are the most efficient due to voxel-based representations and compact temporal encoding. In contrast, 3DGS-based methods achieve higher rendering speed owing to efficient rasterization, with 4DRotorGS attaining the highest FPS.
Overall, 3DGS-based methods offer superior speed but require larger model capacity, while NeRF-style methods are more memory-efficient at the cost of slower rendering.

\begin{table}[tp]
%\rowcolors{2}{gray!15}{white}
\rowcolors{2}{lightcyan}{white}
\caption{\textbf{Performance analysis of 4D reconstruction methods}. GPU memory, frame per second (FPS), and training time are evaluated.}
\label{tab:performance}
\resizebox{\columnwidth}{!}{%
\begin{tabular}{lcccc}
\hline
Methods    & 4D-style           & FPS          & training time (h)  & Params (Mb) \\ \hline
\multicolumn{5}{c}{NeRF-style}                                                     \\ \hline
D-NeRF     & Deformation fields & \textless{}1          & 22.3 & \fst\textbf{3}            \\
DyNeRF     & 4D Primitive       & \textless{}1          & 1344 & \snd7             \\
NeRFPlayer & 4D feature volumes & \textless{}1         & 6    & -                \\
HyperReel  & 4D feature volumes & 6.1           &  2.2         & 360             \\
MixVoxel  & 4D feature volumes & 4.3           &  \trd1.3         & 500             \\
K-Planes  & 4D feature volumes & -             &  3.7         & 51             \\
HexPlanes  & 4D feature volumes & -            &  12          & 200             \\
MSTH  & 4D feature volumes & \fst\textbf{15}                &  \snd0.3         & 135             \\
Ced-NeRF   & 4D feature volumes & \trd6.3           & \fst\textbf{0.2}          & -              \\
StreamRF   & Temporal prior     & \snd10.9         & \snd0.3          & \trd31              \\ \hline
\rowcolor{white} \multicolumn{5}{c}{3DGS-style}                                                     \\ \hline
4DGS    & Explicit 4D Primitive               & 30          & 5.0     & 1183                 \\ 
4DRotorGS   & Explicit 4D Primitive                & \fst\textbf{277}         & 1.0     & -                 \\ 
SpacetimeGS    & Explicit 4D Primitive              & 140          & \snd0.31     & 200                 \\              
CD-3DGS    & Explicit 4D Primitive              & 118          & 1.0     & 338                 \\ 
4D-GS    & Deformation Fields              & 30          & 0.67     & \trd90                 \\ 
ST-4DGS   & Deformation Fields              & 37          & 2.7     & 339                 \\ 
Instant Gaussian Stream    & Deformation Fields               & 204          & \fst\textbf{0.23}     & 2370                 \\ 
GaGS    & Deformation Fields              & 12          & 2.0     & \snd48                   \\ 
DynMF    & Deformation Fields              & 135          & 0.67     & -                 \\ 
HiCoM   & Deformation Fields               & \snd274          & 1.7     & 270                 \\ 
DN-4DGS    & Deformation Fields              & 15          & 0.83     & 112                 \\ 
ED-3DGS    & Deformation Fields              & 74.5          & 1.87     & \fst\textbf{35}                 \\ 
Ex4DGS    & Frame-wise training            & 121          & \trd0.6     & 115   \\
3DGStream    & Frame-wise training              & \trd215          & 1.0     & 2340                 \\ 
4DGC    & Frame-wise training             & 168          & 1.2     & 150  \\ \hline
\end{tabular}%
}
\vspace{-10pt}
\end{table}

%% file: sec/6_4D_future.tex
\section{Future Prospects}
\label{sec:Future}

\subsection{Technical Prospects}

\noindent \textbf{Feed-Forward 4D Representations.}
Existing NeRF and Gaussian Splatting methods largely rely on per-scene optimization, resulting in high computational cost and limited scalability. A major trend is the transition toward generalizable feed-forward models. By leveraging large reconstruction models and transformer-based architectures~\cite{BulletTimer, GS-LRM, DGS-LRM, 4DGT}, the field is gradually shifting from optimizing individual scenes to directly inferring scene representations. This paradigm enables near real-time 4D reconstruction from sparse inputs and helps bridge low-level reconstruction with higher-level spatio-temporal understanding.

\noindent \textbf{Hybrid Explicit--Implicit Representations.}
The distinction between explicit and implicit representations is increasingly converging toward hybrid paradigms. Future 4D models are expected to combine explicit structures for efficient rasterization with implicit latent fields for modeling non-Lambertian effects and temporal topology changes~\cite{spectromotion, NeRF-GS}. Such hybrid designs are important for scaling 4D representations to open-world dynamic scenes, where purely explicit methods face memory limitations and purely implicit methods struggle with real-time interaction.

\noindent \textbf{Integration with Generative Priors.}
Another emerging direction is the integration of generative models~\cite{l4gm,4D-LRM} into 4D reconstruction pipelines. Unlike traditional optimization-based methods that rely heavily on observations, generative priors enable plausible completion, motion prediction, and view synthesis under sparse or degraded inputs~\cite{chen2026unpaired,jiang2026medical}. This integration may shift reconstruction from purely observation-driven modeling toward predictive and generative frameworks, leading to more robust 4D reconstruction and synthesis.

\noindent \textbf{Interactive and Controllable 4D Scenes.}
Beyond passive reconstruction, future 4D systems are expected to support interaction and controllability~\cite{sc-gs}. These capabilities enable users to manipulate dynamic scenes, edit object behaviors, and simulate alternative scenarios. Such developments may transform 4D representations into interactive world models for applications in simulation, digital twins, and content creation.

\subsection{Application Prospects}

\noindent \textbf{Embodied AI Simulation.}
4D reconstruction enables temporally consistent environments for embodied AI~\cite{Robo-GS,Splat-nav, 3dgstream}. Compared with static representations, dynamic 4D scenes provide richer interaction signals and more realistic training environments, supporting robust perception, planning, and interaction in complex settings.

\noindent \textbf{Dynamic Scene Understanding and Editing.}
4D representations facilitate scene understanding, navigation, and editing in dynamic environments~\cite{yan2025renderworld, wang2025unifying, feature4x}. By modeling temporal evolution, these methods support motion prediction, scene simulation, and controllable editing for both analysis and content generation.

\noindent \textbf{Human-Centered Applications.}
4D reconstruction further enables accurate modeling of human motion and interaction~\cite{fan2025rgavatar, zhao2025viewgauss, dai20254d}. These capabilities support applications in AR/VR, telepresence, healthcare, and digital twins. Future systems may further improve realism and interactivity, enabling more immersive user experiences.

%% file: sec/7_conculsion.tex
\section{Conclusion}
\label{sec:Conclusion}

4D scene reconstruction advances spatio-temporal 3D vision by modeling dynamic environments. This survey reviews NeRF- and 3DGS-based approaches, analyzing their efficiency, scalability, and temporal coherence. We further summarize representative datasets, evaluation metrics, and existing methods. Finally, we discuss open challenges and future directions, including feed-forward modeling, hybrid representations, and generative priors, as well as applications in embodied AI and dynamic scene understanding. This survey serves as both a reference and a roadmap for future research in 4D scene reconstruction.

\bibliographystyle{IEEEbib}
\bibliography{ref}